\documentclass[10pt,twocolumn,letterpaper]{article}

\usepackage{iccv}              

\usepackage{multicol}
\usepackage{multirow}
\usepackage{colortbl} 
\usepackage{xcolor}

\usepackage{ifthen}
\usepackage{soul} 
\newboolean{hidecontent} 
\usepackage{marvosym}

\setboolean{hidecontent}{true} 

\usepackage[normalem]{ulem} 

\newcommand{\hideorstrike}[1]{%
    \ifthenelse{\boolean{hidecontent}}{}{%
        \textcolor{red}{\sout{#1}} 
    }%
}

\newcommand{\hideortable}[1]{%
    \ifthenelse{\boolean{hidecontent}}{}{%
        {\color{red} #1} 
    }%
}

\newcommand{\hideorfigure}[2][]{%
    \ifthenelse{\boolean{hidecontent}}{}{%
        \begin{figure}[h]
            \centering
            \includegraphics[#1]{#2}
            \caption{\textcolor{red}{\st{Figure caption}}} 
        \end{figure}
    }%
}
\newcommand{\ptsModify}[1]{\textcolor{black}{#1}}

\definecolor{iccvblue}{rgb}{0.21,0.49,0.74}
\usepackage[pagebackref,breaklinks,colorlinks,allcolors=iccvblue]{hyperref}

\def\paperID{11702} 
\def\confName{ICCV}
\def\confYear{2025}

\definecolor{green}{RGB}{0,150,10}
\definecolor{blue}{RGB}{0,148,181}
\definecolor{orange}{RGB}{194,153,107}

\title{\raisebox{-1.1ex}{\protect\includegraphics[height=2.7\fontcharht\font`\B]{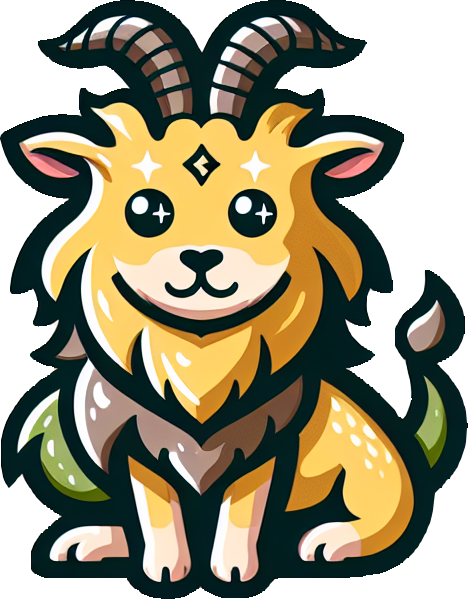}}~Chimera: Improving Generalist Model with Domain-Specific Experts}

\author{
Tianshuo Peng$^{1,2,*}$, Mingsheng Li$^{3,*}$, Jiakang Yuan$^{3}$, Hongbin Zhou$^{1}$, Renqiu Xia$^{1,4}$, Renrui Zhang$^{2}$ \\
Lei Bai$^{1}$, Song Mao$^{1}$, Bin Wang$^{1}$, Aojun Zhou$^{2}$, Botian Shi$^{1}$\\ 
Tao Chen$^{3,5}$, Bo Zhang$^{1,\ddagger,}$\textsuperscript{\Letter}, Xiangyu Yue$^{2,}$\textsuperscript{\Letter}  \\ [1mm]
$^{1}$Shanghai Artificial Intelligence Laboratory, $^{2}$MMLab, The Chinese University of Hong Kong \\
$^{3}$College of Future Information Technology, Fudan University, $^{4}$Shanghai Jiao Tong University \\ $^{5}$Shanghai Innovation Institute \\ [0.5mm]
\normalsize * Equal Contribution, {\normalsize \Letter  \ Corresponding Authors}, \normalsize $\ddagger$~Project Lead \\
}

\begin{document}
\maketitle
\begin{abstract}

%

Large Multi-modal Models (LMMs), trained on web-scale datasets predominantly composed of natural images, have demonstrated remarkable performance on general tasks. 
However, these models often exhibit limited specialized capabilities for domain-specific tasks that require extensive domain prior knowledge. 
An intuitive solution is to post-train LMMs on a specific domain, but often suffers from the labor-intensive annotating process and the inaccessibility of private training data.
Directly integrating expert models tailored for those tasks is also challenging due to representational gaps and imbalanced optimization.
To address these challenges, we introduce \textbf{Chimera}, a scalable and low-cost multi-modal pipeline designed to boost the ability of existing LMMs with domain-specific experts. Specifically, we design a progressive training strategy to integrate features from expert models into the input of a generalist LMM. To address the imbalanced optimization caused by the well-aligned general visual encoder, we introduce a novel Generalist-Specialist Collaboration Masking (GSCM) mechanism. This results in a versatile model that excels across the chart, table, math, and document domains, achieving state-of-the-art performance on multi-modal reasoning and visual content extraction tasks, both of which are challenging tasks for assessing existing LMMs. 
We will release model weights, along with the data used for training and evaluation, to facilitate future research on LMMs.

\end{abstract}    
\vspace{-6pt}
\section{Introduction}
\label{sec:intro}


The past year has witnessed the remarkable success of \textbf{L}arge \textbf{M}ulti-modal \textbf{M}odels (LMMs) in handling a variety of general domain tasks, such as image captioning~\citep{chen2023sharegpt4v,dong2024benchmarking,liu2024playground}, visual dialog~\citep{internvl-1.5,wang2024qwen2,2024claude,openai2023gpt4v,ll3da}, and cross-modal retrieval~\citep{caffagni2024wiki,chartvlm}, demonstrating their potential as a technical pathway towards a general-purpose AI assistant. 
%
{Although proficient in a wide range of tasks, their performance still lags behind that of models fine-tuned with target-domain data, especially in \textbf{specialized tasks} such as \textit{multi-modal reasoning} and \textit{visual content extraction}.
As depicted in Fig.~\ref{fig:compare}, state-of-the-art general-purpose LMMs demonstrate significant limitations in addressing these tasks, highlighting the necessity for further research to bridge this gap.

\begin{figure}[t]
  \centering
   \includegraphics[width=0.97\linewidth]{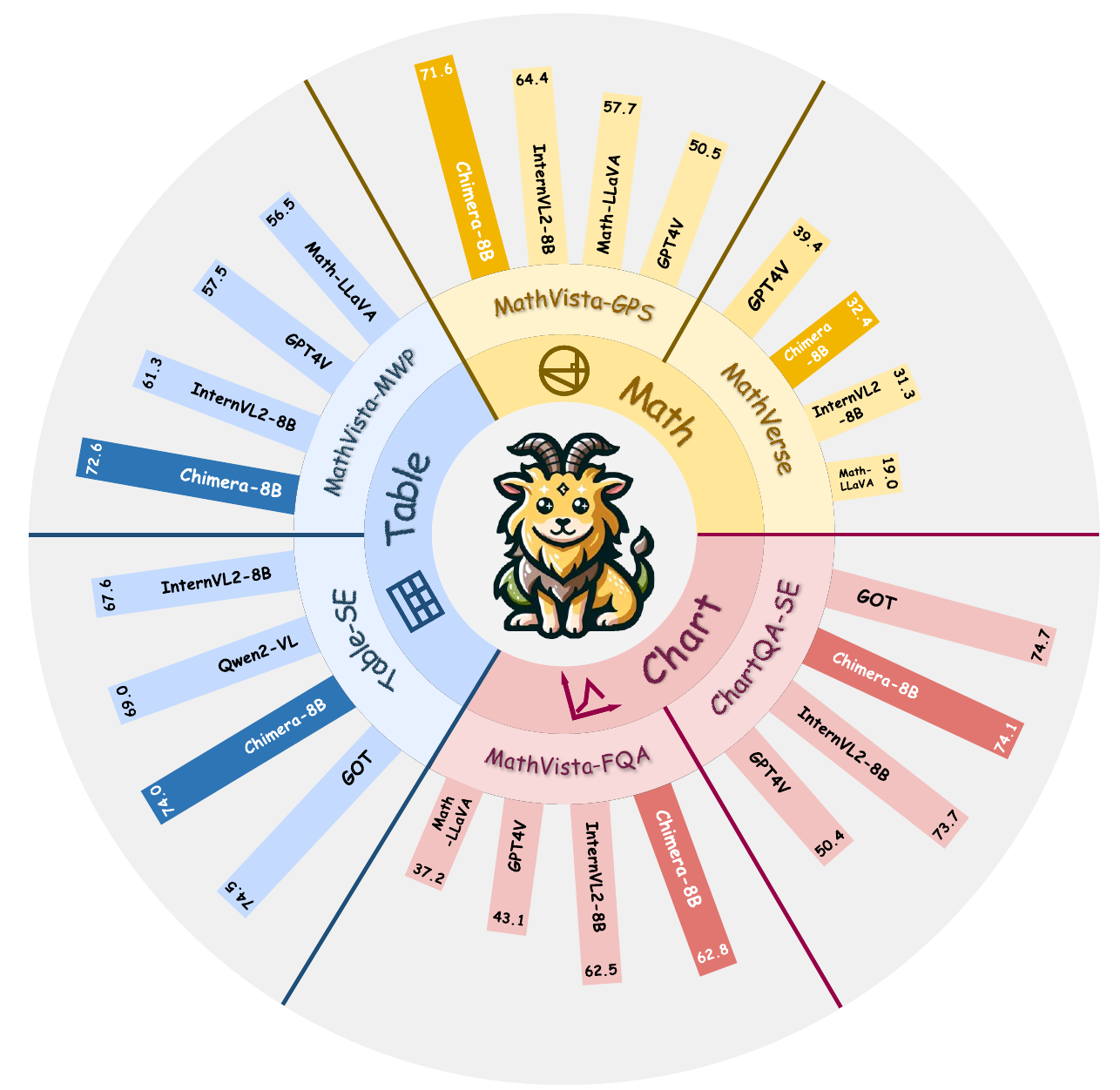}
   \vspace{-10pt}
   \caption{Performance comparison of different models on multi-modal reasoning (MathVista, MathVerse) and visual structural extraction  (ChartQA-SE, Table-SE) tasks.}
   \label{fig:compare}
\vspace{-12pt}
\end{figure}


Current research on LMMs~\citep{internvl-1.5,sphinx,mplugowl-2,openai2023gpt4v,openai2024gpt4o,chen2023sharegpt4v,reid2024gemini} has extensively invested in scaling up by collecting web-scale image-text pairs and employing multi-task instruction tuning to develop generalist models, following a \textit{``One for All"} paradigms~\citep{internvl-1.5,qwen2-vl}.
However, the pursuit of generality often results in suboptimal performance in domain-specific tasks, such as Chart~\citep{chartvlm}, Table~\citep{docgenome}, and Math~\citep{math-puma,jiang2025mme,mavis}. This is mainly due to the substantial differences between natural images and those found in specialized fields~\citep{liang2022mind}.
For instance, domain-specific tasks such as multi-modal reasoning and visual structural extraction often involve content that includes charts, tables, geometric figures, and function graphs~\cite{chartvlm,docgenome,tablellava,gllava}. 
These tasks are characterized by higher text density and more abstract content~\citep{got,wang2024mineru}. 
As a result, general LMMs, which are primarily trained on web-scale natural images, struggle to adapt effectively to these specialized contexts~\citep{llava-onevision,mavis,chartvlm}.


To enhance performance in target domains, numerous studies have focused on developing tailored models or task-specific architectures for downstream tasks~\cite{chartvlm, mavis, mathllava, tablellava, gllava, got, docgenome}, adopting a \textit{``One for One"} paradigm where models are trained on a single scene type. While these expert models exhibit strong capabilities in specialized tasks, they are often criticized for being designed to address individual scenarios. This phenomenon arises from significant distribution gaps across various sub-domains, such as tables, charts, functions, and geometry, potentially sacrificing their generalizability across broader applications using specialized models.

To push the boundary further for the existing LMMs and improve their performance in specialized domains, an intuitive solution is to \textbf{post-train} LMMs on data relevant to the target domain. 
However, a common challenge is that the vast amounts of domain-specific data necessary for specialist models are often proprietary and inaccessible.
On the other hand, integrating specialist experts that contain specialized prior knowledge presents a promising approach to address this issue~\cite{zong2024mova,shen2024mome}.
%
%
Moreover, directly combining specialist experts with the generalist model could result in unsatisfactory performance, due to the following factors: \textbf{1)} large distribution shifts between cross-domain encoders, \textbf{2)} imbalanced optimization for generalists and specialists.

To address these challenges, this work introduces \textbf{Chimera}: a flexible and scalable pipeline that can effectively scale up off-the-shelf experts into LMMs at low cost. 
%
%
Specifically, we utilize a lightweight routing module that dynamically selects tokens from the most suitable experts based on visual content, enabling tailored input to the LLM. Through cost-effective training aimed at feature alignment, we integrate multiple encoders from different expert models into a single LMM, effectively merging diverse specialized knowledge without requiring vast amounts of target-domain data.
Besides, we observed alignment imbalances during the cross-modal encoder fusion and propose a General-Expert Collaboration Masking mechanism to facilitate better model fusion.
Our method easily adapts LMMs, such as InternVL~\citep{internvl-1.5,chen2024internvl}, to a range of domain-specific tasks, including advanced mathematical reasoning, table/chart QA \& extraction, and document structural extraction tasks. By aggregating multiple expert models into a single general LMM, Chimera develops a versatile model endowed with multiple specialized capabilities. 
During inference, Chimera employs a simple routing module to determine whether to invoke the corresponding domain expert model based on the visual input, resulting in a versatile model that excels across the chart, table, math, and document domains, as well as tasks involving multi-modal reasoning and extraction.



We conduct extensive experiments to evaluate Chimera's capabilities in multi-modal reasoning and visual content extraction, both of which are challenging domains for assessing existing LMMs.
With the introduction of domain knowledge from expert models and supervised fine-tuning, Chimera achieves overall accuracies of 64.9 and 32.4 on the multi-modal reasoning benchmarks MathVista~\citep{lu2023mathvista} and MathVerse~\citep{zhang2025mathverse}, setting a new State-Of-The-Art (SOTA) for LMMs of comparable scale.
Direct preference optimization can further boost Chimera's reasoning capabilities, allowing it to achieve superior performance with a small amount of data.
It also surpasses or matches the performance of representative expert models in visual content extraction tasks across chart, table, and document domains.

Our contributions can be summarized as follows:
\begin{enumerate}
\item We introduce Chimera, a scalable pipeline that integrates specialist models into generalist LMMs, facilitating their adaptation to many specialized tasks.
\item We present a lightweight routing module that dynamically selects the most relevant experts based on visual input, coupled with Generalist-Specialist Collaboration Masking (GSCM) aimed at facilitating representation alignment between the generalist and domain experts.
\item Chimera achieves SOTA performance on challenging benchmarks for reasoning, including MathVista and MathVerse. Furthermore, it achieves near-specialist-level results in visual structural extraction on benchmarks like ChartQA-SE, Table-SE, Doc-SE, \textit{etc}.
\end{enumerate}

\vspace{-4pt}
\section{Related Work}
\label{sec:related}

\noindent \textbf{Generalist Large Multi-modal Models.} Following the remarkable success of Large Language Models (LLMs)~\citep{gpt3/brown2020language,llama/touvron2023llama,llama2/touvron2023llama}, researchers have made great efforts in adapting LLMs for multi-modal tasks in a general context, contributing to the flourishing of Large Multi-modal Models (LMMs)~\citep{internvl-1.5,sphinx,mplugowl-2,openai2023gpt4v,openai2024gpt4o,chen2023sharegpt4v,reid2024gemini,m3dbench,lu2025omnicaptioner}.
Recent LMMs typically utilize a cross-modal connector~\citep{internvl-1.5,blip-2} and perform the pre-training on large-scale natural image-text datasets~\citep{li2024omnicorpus,docgenome} to alleviate the modality gap between the visual encoder and the LLMs.
For instance, BLIP series~\cite{blip, blip-2} utilizes captions from datasets like COCO~\cite{coco}, CC3M~\cite{cc3m}, SBU~\cite{sbu}, and LAION~\cite{laion}, while the LLaVA series~\cite{llava-1, llava-1.5} constructs complex instruction-following datasets based on natural images from COCO~\cite{coco} to further enhance their understanding of visual content. 

However, the pursuit of generality often results in limited performance in specialized scenarios, such as geometric and function reasoning~\cite{xia2024geox,gllava,mavis,mme-reasoning}, table and chart understanding~\cite{chartvlm,tablellava,wang2024mineru}, of which the visual content differing differs significantly from natural images. 
Moreover, fine-tuning LMMs on specialized domains remains challenging due to inaccessible private data and potential degradation in general performance.



\begin{figure*}[t]
    \centering
    \includegraphics[width=0.9\textwidth]{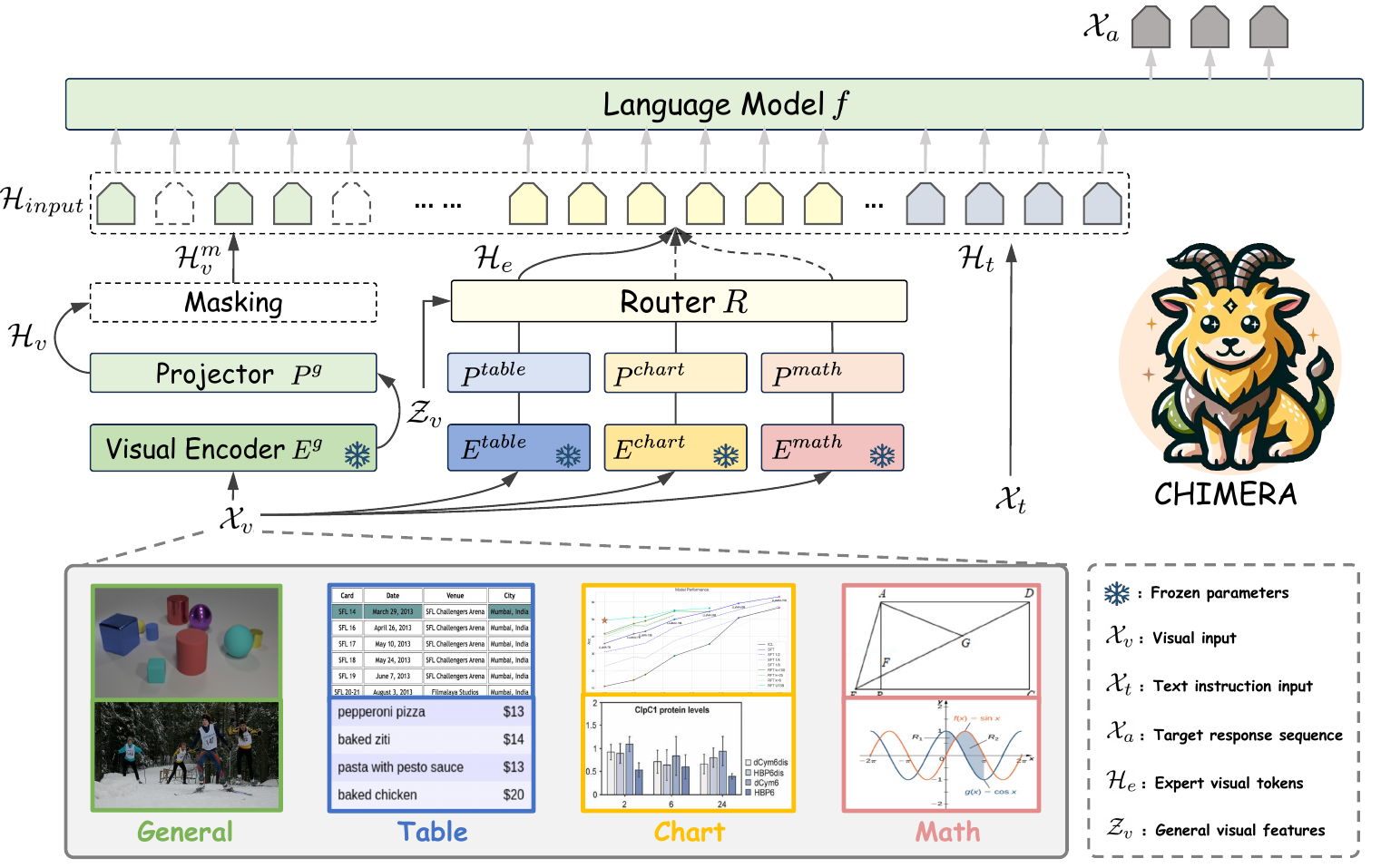}
    \vspace{-4pt}
    \caption{\textbf{Overview of our Chimera framework}. Chimera uses Generalist-Specialist Collaboration Masking to facilitate the alignment with expert models. During inference, the Router $R$ decides expert invocations based on the visual input, resulting in a versatile model that excels across multiple specialized domains and tasks.
    }
    \label{fig:overview}
\vspace{-8pt}
\end{figure*}

\noindent \textbf{Expert Models on Specialized Scenarios.}
Expert tasks in multimodal settings, such as geometric and function reasoning~\cite{xia2024geox,zhang2025mathverse,mavis}, table and chart understanding~\cite{tablellava,chartvlm}, and document information extraction~\cite{wang2024mineru}, often require specialized designs to achieve optimal task performance. For example, Math-LLaVA~\cite{mathllava}, and MAVIS~\cite{mavis} train LMMs on carefully curated mathematical datasets using natural language descriptions. Table-LLaVA~\cite{tablellava} constructs a large-scale multimodal table understanding dataset, while StructEqTable~\cite{docgenome} uses extensive table format transformation data to build a highly specialized expert model with limited generality. Similarly, ChartGemma~\cite{chartgemma} and ChartInstruct~\cite{chartinstruct} train LMMs on diverse chart instruction-following data, and ChartVLM~\cite{chartvlm} employs a router structure to selectively engage different decoders for base perception tasks and cognition tasks. GOT~\cite{got} trains a specialist model on million-scale private data specifically for document structural extraction task.
Besides, some work executes complex and challenging tasks, including automated scientific discovery, in the form of multi-agents through carefully designed and comprehensive workflows~\cite{team2025novelseek}.



While expert models excel in specific domains, they struggle with tasks outside their specialization. In contrast, Chimera integrates specialized knowledge into a generalist LMM, achieving superior performance across both multi-modal reasoning and document context extraction tasks.

\section{Methodology}
\label{sec:method}
\vspace{-4pt}


To develop an assistant that can adapt to challenging domains at a low cost, we propose Chimera, a scalable multi-modal model built upon existing LMMs and pretrained expert models.
In this section, we first introduce an overview of Chimera in~\cref{subsec:overview}. \cref{subsec:integration} discusses the integration of generalists and domain-specific experts, and \cref{subsec:gscm} details 
Generalist-Specialist Collaboration Masking (GSCM), an effective method for collaboration between the generalist model and specialists.
Furthermore, we present the training recipe for Chimera in \cref{sec:training recipe}.


\vspace{-4pt}
\subsection{Overview}
\label{subsec:overview}
\vspace{-4pt}


As illustrated in Fig~\ref{fig:overview}, Chimera consists of: a general visual encoder $E^{g}$, a general projector $P^{g}$ together with a language Model $f$ initialized from a pretrained LMM, a router $R$, an expert model set $S^{e}$ with $N_e$ expert models and corresponding expert projector set $S^{p}$.
Assuming expert models from the domains of table, chart, and math as aggregation targets, we have: 
\begin{equation}
\begin{aligned}
&S^{e} = \{E^{table}, E^{chart}, E^{math}\} \\
&S^{p} = \{P^{table}, P^{chart}, P^{math}\}.
\end{aligned}
\end{equation}

\noindent \textbf{Generalist Branch.} For visual input $\mathcal{X}_{v}$, $E^{g}$ privides the general visual features $\mathcal{Z}_v=E^{g}(\mathcal{X}_{v})$, $P^{g}$ projects general visual features into word embedding space, yielding general visual tokens $\mathcal{H}_v=P^{g}(\mathcal{Z}_{v})$.
During training, we apply the GSCM mechanism on $\mathcal{H}_v$ as $\mathcal{H}_v^{m}$ to replace $\mathcal{H}_v$.

\noindent \textbf{Specialist Branch.} The linear layer $R$ first predicts routing value $\mathcal{H}_r\in R^{N_e+1}$ as $\mathcal{H}_r = R(\mathcal{Z}_{v}^{cls})$, where $\mathcal{Z}_{v}^{cls}$ represents the classification token of $\mathcal{Z}_{v}$, determining whether to invoke an expert model and which specific expert model to call. Consequently, the expert visual tokens $\mathcal{H}_e$ can be formulated as:
\begin{equation}
\begin{aligned}
i &= \arg\max_i(\mathcal{H}_r)_i \\
\mathcal{H}_e &= \left\{
\begin{matrix}
\varnothing, & \text{if } i==0, \\
S^{p}_{i}\left(S^{e}_{i}\left(\mathcal{X}_{v}\right)\right), & \text{otherwise}.
\end{matrix}
\right.  
\end{aligned}
\end{equation}


Given the text embedding $\mathcal{H}_{t}$ of instruction $\mathcal{X}_{t}$, the input sequence during training is formulated as:
\begin{equation}
\begin{aligned}
    &\mathcal{H}_{input}  = \text{concat}([\mathcal{H}_v^{m}:\mathcal{H}_e:\mathcal{H}_t]). \\
\end{aligned}
\end{equation}

\hideorstrike{
We construct two variants of Chimera: \textit{Chimera-Reasoner} and \textit{Chimera-Extractor}. \textit{Chimera-Reasoner} aims to integrate expert models specializing in table, chart and math domains, making it well-suited for comprehensive multi-modal reasoning scenarios. \textit{Chimera-Extractor} focuses on integrating expert models that specialize in document structural extraction. This task is more specialized and involves richer visual-text information.
}

We validated Chimera's capability and adaptability in two distinctly different task scenarios: multi-modal reasoning and visual content extraction, both of which are challenging domains for assessing existing LMMs. 
The former scenario requires integrating expert models including those for tables, math, and charts. The latter scenario requires integrating expert models that specialize in document structural extraction.



\vspace{-4pt}
\subsection{Integration of Generalist and Specialist}
\label{subsec:integration}
\vspace{-4pt}

There are two intuitive ideas to adapt an generalist LMM into specialized domain: \textbf{1)} performing supervised fine-tuning on domain-specific data (\underline{naive finetune}) and \textbf{2)} sequentially appending features from different encoders (\underline{naive concat}).
The primary difference between Chimera and these two approaches lies in the definition of the input sequence during training for the language model $f$. Let $\mathcal{H}_{input}^{nf}$ and $\mathcal{H}_{input}^{nc}$ denote the input sequences in the naive finetune and naive concat methods, respectively. They can be formulated as:
\begin{equation}
\begin{aligned}
    &\mathcal{H}_{input}^{nf}  = \text{concat}([\mathcal{H}_v:\mathcal{H}_t]), \\
    &\mathcal{H}_{input}^{nc}  = \text{concat}([\mathcal{H}_v:\mathcal{H}_e:\mathcal{H}_t]). \\
\end{aligned}
\end{equation}

The former approach attempts to use a single visual encoder to handle all visual content, which refuses to incorporate domain-specific knowledge from expert models. Fine-tuning on subtasks in several specialized domains can also reduce generalizability, leading to trade-offs or suboptimal performance across subtasks.
The latter approach incorporates encoded features from various domains, but applying this directly to a well-aligned LMM may lead to misalignment between the generalist and specialist models—a limitation we will discuss in the next section.

\subsection{Generalist-Specialist Collaboration Masking}

\label{subsec:gscm}

Although naive concat method with input $\mathcal{H}_{input}^{nc}$ is intuitive, we still concern that since the general visual encoder $E^{g}$ is well-aligned with language Model $f$, it may cause the model to overly rely on $E^{g}$ to complete tasks, which leads to ineffective alignment with the expert models.
To better align domain knowledge and general world knowledge, we propose a simple yet effective learning mechanism called Generalist-Specialist Collaboration Masking, designed to boost the synergy between general-purpose and domain-specific capabilities.

During training, we sample a subset of general visual tokens from $\mathcal{H}_v$ at a certain ratio and mask them to build the masked general visual tokens $\mathcal{H}_v^m$.
In practice, this is achieved by setting the attention mask corresponding to the sampled subset to False.
We consider a simple sampling strategy: randomly sampling tokens without replacement according to a uniform distribution. 
Applying mask to information provided by general encoder $E^{g}$ produces a limitation on $E^{g}$, which will force the model to utilize domain-specific information provided by expert models as supplements for vision-language tasks.
The uniform distribution helps prevent bias that may arise from masking predominantly in the image center or specific regions.

\subsection{Training Recipe}

\label{sec:training recipe}


To equip multi-modal generalists with rich domain-specific knowledge, we apply a progressive training strategy, including Domain-General Knowledge Alignment and Visual Instruction Tuning. 
\hideorstrike{
The dataset used during training can be found in Tab.~\ref{tab:reasoner_data} and Tab.~\ref{tab:extractor_data}. 
Through two-stage training, we develop two versatile models: \textit{Chimera-Reasoner} and \textit{Chimera-Extractor}. The Chimera-Reasoner takes questions and images as inputs, performing reasoning and answering, while the Chimera-Extractor excels at extracting structured information from visual documents.
}
Through two-stage training, we develop variants for different scenarios respectively. 
All the datasets used are publicly available. Details about the datasets can be found in the supplementary material.

\noindent \textbf{Domain-General Knowledge Alignment.}~To initially align domain-specific knowledge with the semantic space of the generalist LMM, we train the model using tasks that directly perceive diverse image content. 
The tasks include natural image description, table format transformation, chart structural extraction and summarization, math diagram captioning, and paragraph-level OCR.

With guidance from image-text pairs across different domains, the model is able to leverage domain knowledge from expert models to accurately recognize visual content in each domain and describe its spatial arrangement. This marks the first step toward deeper integration. In this stage, we freeze the general visual encoder $E^{g}$, expert model set $S^{e}$ and language model $f$, only train the router $R$, general projector $P^{g}$ and expert projector set $S^{p}$.

\noindent \textbf{Visual Instruction Tuning.} To further align model with domain knowledge from expert models and enhance its performance on specialized tasks across different domains, we take instruction-following datasets from various domains to perform visual instruction tuning with the proposed GSCM.
%
During this stage, we unfreeze router $R$, general projector $P^{g}$, expert projector set $S^{p}$ and language model $f$, perform a thorough instruction-following tuning, which finally results in the versatile Chimera models.

\noindent \textbf{Training Objective.}~Our primary training objective is to optimize the trainable parameters $\theta$, so that the likelihood of target response sequence $\mathcal{X}_{a}$ is maximized given the visual input $\mathcal{X}_{v}$ and instruction $\mathcal{X}_{t}$ as follows:
\begin{equation}
    \theta^{*} = \arg \max_{\theta} P(\mathcal{X}_{a} \vert \mathcal{X}_{v}, \mathcal{X}_{t}; \theta).
\end{equation}


To accomplish this, we utilize token-wise cross-entropy loss to train the model in an auto-regressive manner. For target $\mathcal{X}_a$ of length $L$, the auto-regressive modeling loss $\mathcal{L}_m$ is represented as follows:
\begin{equation}
    \mathcal{L}_m = -\sum_{i=1}^{L} \log P(x_i |\mathcal{X}_{v},  \mathcal{X}_{t}, \mathcal{X}_{a;<i}, \theta),
\end{equation}
where $\mathcal{X}_{a;<i}$ are the tokens before the current prediction token $x_i$.

Besides, we add classification loss to guide the Router to accurately call different expert models based on image content,which can be represented as follows:
\begin{equation}
    \mathcal{L}_c = -\sum_{i=0}^{N_e+1} \log P(c_i |\mathcal{X}_{v}, \theta),
\end{equation}
where $c_i$ represents the expert domain category that the current image requires for invocation (including category $0$, which means no expert model is invoked). Finally, the optimization objective is formulated as follows:
\begin{equation}
    \mathcal{L} = \mathcal{L}_c + \mathcal{L}_m.
\end{equation}

\vspace{-6pt}
\section{Experiments}
\vspace{-4pt}

\label{sec:experiments}

\begin{table*}[ht!]
\vspace{-10pt}
\centering
\renewcommand\tabcolsep{3pt} 
\renewcommand\arraystretch{1.05} 
\tiny 
\resizebox{0.98\textwidth}{!}{
\begin{tabular}{l|c|c|ccccc|ccccccc}
\toprule
Model & \#Params. & ALL & FQA & GPS & MWP & TQA & VQA & ALG & ARI & GEO & LOG & NUM & SCI & STA \\ 
\midrule
\rowcolor{gray!20}
\multicolumn{15}{c}{\textit{\textbf{Close Source LMMs}}} \\
InternVL2-Pro~\citep{internvl-1.5}& - & 66.8 & 70.6 & 65.4 & 76.9 & 71.5 & 48.0 & 66.5 & 62.3 & 63.6 & 27.0 & 40.3 & 65.6 & 81.1 \\
Gemini 1.5 Pro~\citep{team2024gemini}& - & 63.9 & - & - & - & - & - & - & - & - & - & - & - & - \\
GPT-4o& - & 63.8 & - & - & - & - & - & - & - & - & - & - & - & - \\
Grok-1.5V& - & 52.8 & - & - & - & - & - & - & - & - & - & - & - & - \\
Claude 3 Opus~\citep{claude3}& - & 50.5 & - & - & - & - & - & - & - & - & - & - & - & - \\
GPT-4V (Playground)& - & 49.9 & 43.1 & 50.5 & 57.5 & 65.2 & 38.0 & 53.0 & 49.0 & 51.0 & 21.6 & 20.1 & 63.1 & 55.8 \\
\midrule
\rowcolor{gray!20}
\multicolumn{15}{c}{\textit{\textbf{Open Source LMMs}}} \\
LLaVA-OneVision~\citep{llava-onevision}&72B & 67.5 & - & - & - & - & - & - & - & - & - & - & - & -   \\
Math-LLaVA$^*$~\cite{mathllava}&13B & 46.6 & 37.2 & 57.7 & 56.5 & 51.3 & 33.5 & 53.0 & 40.2 & 56.5 & 16.2 & 33.3 & 49.2 & 43.9 \\
Pixtral~\citep{agrawal2024pixtral} & 12B & 58.0 & - & - & - & - & - & - & - & - & - & - & - & - \\
SPHINX-MoE~\citep{sphinx}& 8×7B &  42.7 & - & - & - & - & - & - & - & - & - & - & - & -   \\
InternLM-XComposer2~\citep{ixc-2}& 7B & 57.6 & 55.0 & 63.0 & \textbf{73.7} & 56.3 & 39.7 & 56.6 & 52.4 & 62.3 & 8.1 & 42.4 & 59.0 & 64.1 \\
LLaVA-OneVision~\cite{llava-onevision}&7B & 63.2 & - & - & - & - & - & - & - & - & - & - & - & -   \\
Math-PUMA-DeepSeek-Math$^*$~\cite{math-puma} & 7B & 44.7 & 42.8 & 39.9 &67.7 & 42.4 &31.3 & 39.2 &41.9 &41.4 &8.1 & 36.8 & 48.4 &52.5  \\
\midrule
\multirow{2}{*}{Qwen2-VL~\cite{qwen2-vl}}&2B & 43.0 & - & - & - & - & - & - & - & - & - & - & - & -   \\
&7B & 58.2 & - & - & - & - & - & - & - & - & - & - & - & -   \\
\midrule
\multirow{3}{*}{IntenrVL2~\cite{internvl-1.5}} & 2B & 48.3 & 51.3 & 45.7 & 40.9 & 50.6 & 52.5 & 43.4 & 47.3 & 42.3 & 13.5 & 28.5 & 53.3 & 56.8 \\ 
 & 4B & 57.0 & 58.0 & 58.2 & 62.4 & 57.0 & 48.6 & 55.9 & 53.8 & 55.2 & 13.5 & 30.6 & 59.0 & 65.1 \\ 
 & 8B & 61.6 & 62.5 & 64.4 & 61.3 & 64.6 & \textbf{54.7} & 63.0 & \textbf{58.9} & 61.9 & 18.9 & 34.0 & 59.0 & 70.1 \\ 
\midrule
\multirow{3}{*}{Chimera}  & 2B & 53.1 & 52.4 & 56.7 & 62.9 & 51.9 & 40.8 & 52.7 & 47.6 & 56.1 & 10.8 & 34.0 & 52.5 & 61.1 \\ 
& 4B & 61.3 & 58.4 & 66.8 & 72.0 & 61.4 & 48.0 & 63.3 & 54.7 & 65.7 & \textbf{24.3} & 39.6 & 60.7 & 66.4 \\ 
& 8B & \textbf{64.9} & \textbf{62.8} & \textbf{71.6} & 72.6 & \textbf{65.2} & 52.0 & \textbf{67.6} & 57.8 & \textbf{69.5} & 21.6 & \textbf{45.8} & \textbf{61.5} & \textbf{69.4} \\ 
\midrule
\rowcolor{gray!20}
\multicolumn{15}{c}{\textit{\textbf{LMMs with Preference Optimization}}} \\
InternVL-MPO~\cite{wang2024mpo} & 8B & 67.3 &\textbf{72.5} &73.6 &69.9 &\textbf{66.5} &50.3 & 70.1 & 57.5 & 71.5 & 27.0 & 43.1 & 65.6 & 79.1\\
Ovis1.6-Gemma2~\cite{lu2024ovis} & 9B & 67.2 & - & - & - & - & -  & - & - & - & - & - & - & - \\
Chimera$^\dagger$ & 8B &\textbf{68.3} & 66.5 & \textbf{76.9} & \textbf{80.1} & 60.8 & \textbf{55.3} & \textbf{69.7} & \textbf{64.8} & \textbf{74.5} & 13.5 & \textbf{49.3} & \textbf{62.2} &\textbf{77.7}\\
\midrule
Human performance&- & 60.3 & 59.7 & 48.4 & 73.0 & 63.2 & 55.9 & 50.9 & 59.2 & 51.4 & 40.7 & 53.8 & 64.9 & 63.9 \\
\bottomrule
\end{tabular}
}
\caption{Accuracy scores on the \textit{testmini} subset of MathVista. Task types: FQA: figure QA, GPS: geometry problem solving, MWP: math word problem, TQA: textbook QA, VQA: visual QA.
Math reasoning types: ALG: algebraic, ARI: arithmetic, GEO: geometry, LOG: logical , NUM: numeric, SCI: scientific, STA: statistical.
Chimera$^\dagger$ represents post-trained Chimera model.
$^*$ represents the domain expert model.
Chimera set a new SOTA results of 64.9 among open source LMMs under 70B scale. Direct preference optimization further boost Chimera's performance to 68.3, outperforming the latest LMMs with preference optimization.
}
\label{tab:mathvista}
\vspace{-8pt}
\end{table*}

\begin{table*}[!t]
\centering
\renewcommand\tabcolsep{3pt} 
\renewcommand\arraystretch{1.05} 
\tiny
\resizebox{\textwidth}{!}{
\begin{tabular}{l|c|c|cccccc}
\toprule
Model & \#Params. & All Acc & Text Dominant & Text Lite & Vision Intensive & Vision Dominant & Vision Only \\ 
\midrule
\rowcolor{gray!20}
\multicolumn{8}{c}{\textit{\textbf{Closed-source MLLMs}}} \\
Gemini-Pro~\citep{team2023gemini} & - & 23.5 & 26.3 & 23.5 & 23.0 & 22.3 & 22.2 \\
Qwen-VL-Max~\citep{qwen-vl} & - & 25.3 & 30.7 & 26.1 & 24.1 & 24.1 & 21.4 \\
GPT-4V & - & 39.4 & 54.7 & 41.4 & 34.9 & 34.4 & 31.6 \\
\midrule
\rowcolor{gray!20}
\multicolumn{8}{c}{\textit{\textbf{Open-source MLLMs}}} \\
SPHINX-Plus~\citep{sphinx} & 13B & 14.0 & 16.3 & 12.8 & 12.9 & 14.7 & 13.2 \\
SPHINX-MoE~\citep{sphinx} & 8×7B & 15.0 & 22.2 & 16.4 & 14.8 & 12.6 & 9.1 \\
LLaVA-NeXT~\citep{li2024llavanext-strong} & 110B & 24.5 & 31.7 & 24.1 & 24.0 & 22.1 & \textbf{20.7} \\
LLaVA-NeXT~\citep{li2024llavanext-strong} & 8B & 19.3 & 24.9 & 20.9 & 20.8 & 16.1 & 13.8 \\
InternLM-XComposer2~\citep{ixc-2} & 7B & 16.5 & 22.3 & 17.0 & 15.7 & 16.4 & 11.0 \\
Math-LLaVA$^*$~\citep{mathllava} & 13B & 19.0 & 21.2 & 19.8 & 20.2 & 17.6 & 16.4 \\
MAVIS-7B$^*$~\citep{mavis} & 7B & 27.5 & 41.4 & 29.1 & 27.4 & 24.9 & 14.6 \\
Math-PUMA-DeepSeek-Math$^*$~\cite{math-puma} & 7B & 31.8 & \textbf{43.4} & 35.4 & 33.6 &31.6 &14.7 \\
\midrule
\multirow{3}{*}{InternVL2~\citep{internvl-1.5}} & 2B & 21.4 & 24.1 & 22.5 & 22.8 & 21.1 & 16.6 \\ 
 & 4B & 26.3 & 32.0 & 28.6 & 28.0 & 24.4 & 18.8 \\ 
 & 8B & 31.3 & 38.8 & 34.5 & 33.6 & 32.6 & 17.0 \\ 
\midrule
\multirow{3}{*}{Chimera}  & 2B & 22.6 & 27.3 & 23.9 & 22.3 & 22.8 & 16.9 \\ 
 & 4B & 27.2 & 31.4 & 30.8 & 29.7 & 25.7 & 18.2 \\ 
 & 8B & \textbf{32.4} & 39.6 & \textbf{35.8} & \textbf{34.8} & \textbf{32.7} & 19.3 \\ 
\bottomrule
\end{tabular}
}
\vspace{-4pt}
\caption{Performance Comparison on MathVerse with the accuracy metric.
$^*$ represents the domain expert model.}
\vspace{-8pt}
\label{tab:mathverse}
\end{table*}

\begin{table}[ht!]
\centering
\renewcommand\tabcolsep{3pt} 
\renewcommand\arraystretch{1.0} 
\vspace{4pt}
\tiny 
\resizebox{1.0\columnwidth}{!}{
\begin{tabular}{c|c|cccc}
\toprule
Model & ALL & General &Chart &Table &Math \\ 
\midrule
InternVL2-2B~\citep{internvl-1.5} & 48.3 & 45.3 & 58.9 & 50.0 & 44.2 \\
InternVL2-4B~\citep{internvl-1.5} & 57.0 & 50.1 & 66.2 & 65.7 & 58.3 \\ 
InternVL2-8B~\citep{internvl-1.5} & 61.6 & 52.7 & \textbf{71.2} & 67.1 & 66.5 \\ 
Chimera-2B & 53.1 & 46.0 & 60.3 & 62.9 & 56.1 \\
Chimera-4B & 61.3 & 54.0 & 64.8 & \textbf{72.9} & 66.9 \\ 
Chimera-8B & \textbf{64.9} & \textbf{57.5} & \textbf{71.2} & 62.9 & \textbf{71.9} \\ 
\bottomrule
\end{tabular}
}
\vspace{-4pt}
\caption{Accuracy scores of different visual content domain on the \textit{testmini} subset of MathVista.Those do not belong to the last three domains are uniformly classified as General for simplicity.}
\label{tab:mathvista_domain}
\vspace{-10pt}
\end{table}
\begin{table*}[!h]
\centering

\resizebox{\textwidth}{!}{
\begin{tabular}{cccccccccc}
\toprule
Task & Metric & Deplot$^*$~\citep{liu2022deplot} & UniChart$^*$~\citep{masry2023unichart} & ChartVLM$^*$~\citep{chartvlm} & GPT-4V & Qwen-VL~\citep{qwen-vl} & GOT~\citep{got} & InternVL-2~\cite{internvl-1.5}  & Chimera \\ 
\midrule
\multirow{3}{*}{ChartQA-SE}
& AP@strict  & 61.4  & 42.3 & 71.8 & 50.4 &  58.6   &  \textbf{74.7} & 73.7 & 74.1 \\ 
& AP@slight  & 70.9 & 53.1 & 81.4 & 60.6 & 68.5   &  \textbf{84.5} & 83.9 & 84.4 \\ 
& AP@high    & 72.9 & 56.0  & 84.2 & 64.3 &  72.7   &  86.7 & 87.2 & \textbf{87.6} \\ 
\midrule
\multirow{3}{*}{PlotQA-SE}
& AP@strict  & 3.1 & 10.5 & 3.8 & 7.3 &  0.5   & \textbf{13.3} & 5.7 & 5.9 \\ 
& AP@slight  & 16.5 & 26.0 & 46.8 & 19.4 & 4.2    & 59.6 & 55.0 & \textbf{62.1} \\ 
& AP@high    & 26.5 & 26.9  & 54.0 & 22.3 &  12.0   & 64.0 & 61.8 & \textbf{71.0} \\ 
\bottomrule
\end{tabular}
}
\vspace{-4pt}
\caption{Performance on ChartQA-SE and PlotQA-SE. Metrics include Average Precision (AP) at strict, slight, and high levels.
$^*$ represents the domain expert model.}

\label{tab:chartse}
\vspace{-4pt}
\end{table*}

\begin{table}[t]
\begin{minipage}{0.5\textwidth}
\centering
\resizebox{\columnwidth}{!}{
\begin{tabular}{cccc}
\toprule
Method & Edit Distance$\downarrow$  & TEDS$\uparrow$ & TEDS (structure only)$\uparrow$   \\  
\midrule
InternVL-2~\citep{internvl-1.5} & 0.229 & 0.676 & 0.762  \\
Qwen2-VL~\citep{qwen2-vl} & 0.231 & 0.690 & 0.773  \\
StructEqTable$^*$~\citep{docgenome} & 0.226 & 0.706 & 0.787  \\
GOT$^*$~\citep{got} & 0.257& \textbf{0.745} & \textbf{0.830}  \\
Chimera & \textbf{0.165} & 0.740 & 0.828  \\
\bottomrule
\end{tabular}
}
\vspace{-4pt}
\caption{Comparison of performance on Table-SE across different methods: TEDS, TEDS (structure only), and Edit Distance.
$^*$ represents the domain expert model.}
\label{tab:tablese}
\end{minipage}
\end{table}




\begin{table}[t]
\vspace{-2pt}
\begin{minipage}{0.5\textwidth}
\centering
\resizebox{1.0\textwidth}{!}{
\begin{tabular}{ccccccccc}
\toprule
\multirow{2}{*}{{Method}} & \multicolumn{2}{c}{Edit Distance$\downarrow$} & \multicolumn{2}{c}{Precision$\uparrow$}   & \multicolumn{2}{c}{BLEU$\uparrow$} & \multicolumn{2}{c}{METEOR$\uparrow$}  \\  
\cmidrule(rl){2-3} \cmidrule(rl){4-5}  \cmidrule(rl){6-7}  \cmidrule(rl){8-9}  
& en & zh  & en & zh  & en & zh & en & zh \\ 
\midrule
InternVL~\citep{internvl-1.5} & 0.504 & 0.604 & 65.4 &66.0  & 38.4 & 33.1 & 52.6 & 50.6 \\
GOT$^*$~\citep{got} & 0.355 & 0.510 & 67.9& \textbf{71.2} & \textbf{52.5} & 34.3 & \textbf{65.3} & 53.9  \\
Chimera & \textbf{0.304} & \textbf{0.461} &\textbf{69.6} & 66.1 & 49.8 & \textbf{40.5} & 64.8 & \textbf{56.9} \\
\bottomrule
\end{tabular}
}
\vspace{-4pt}
\caption{Comparison of performance metrics across different methods on Doc-SE. Metrics include Edit Distance (lower is better), Precision, BLEU, and METEOR (higher is better).
$^*$ represents the domain expert model.}
\label{tab:inhouse-page}
\end{minipage}

\end{table}

To evaluate the capabilities of Chimera, we begin by detailing the datasets and metrics used for evaluation, along with the implementations for multi-modal reasoning and visual content extraction in~\cref{subsec:implementation}. Then, we provide a comparison of Chimera models against previous generalist models across various benchmarks, including multi-modal reasoning~(\cref{subsec:multi-modal reasoning}) and visual structural extraction~(\cref{subsec:visual content extraction}).
\hideorstrike{
Besides, we conduct quantitative ablation studies on the model design and training strategy in~\cref{subsec:ablation}.
}
Besides, we conduct further analysis on the model design and training strategy in~\cref{subsec:further_analysis}.

%
%
%

\vspace{-4pt}
\subsection{Datasets, Metrics and Implementation Details}
\label{subsec:implementation}
\vspace{-4pt}

\noindent \textbf{Datasets.} In this paper, we conduct quantitative evaluations of Chimera model across a range of challenging multi-modal benchmarks, categorized into the following areas:
\begin{itemize}
    \item \textbf{Multi-modal Reasoning.} 
    We evaluate the Chimera on the MathVista~\cite{lu2023mathvista} to determine its visual reasoning capabilities. 
    Besides, we extend our evaluation on the MathVerse~\cite{zhang2025mathverse}, which is specifically designed for mathematical problem-solving, to gauge its performance in multi-modal mathematical reasoning.
    \item \textbf{Visual Structural Extraction (SE).} 
    Evaluations of Chimera on chart domain are conducted on the challenging ChartQA-SE~\cite{masry2022chartqa} and PlotQA-SE~\cite{chartvlm} benchmarks. 
    Following the protocol of StructChart~\cite{xia2023structchart}, we utilize the test sets of both the ChartQA~\cite{masry2022chartqa} and PlotQA~\cite{methani2020plotqa} to ensure a fair comparison. 
    For table and document, we manually collected and annotated a table format transformation benchmark called Table-SE and a document structural extraction benchmark called Doc-SE. Details regarding the data collection and annotation process can be found in the supplementary material.
\end{itemize}


\noindent \textbf{Metrics.} In evaluations, we adhere to the default metrics used by benchmarks, such as MathVista, MathVerse, ChartQA-SE, and PlotQA-SE. 
For the assessment of the table format transformation, we use Tree-EditDistance-based Similarity (TEDS) score
and Edit Distance for evaluation.
For document structural extraction, we take Edit Distance, Precision, BLEU and METEOR as evaluation metrics.


\noindent \textbf{Implementation Details.} We initialize Chimera using the InternVL2 series. 
Specifically, we use InternVL2-2B, 4B, 8B to construct Chimera-2B, 4B, and 8B for multi-modal reasoning, while using InternVL-1B to build Chimera for visual content extraction.
In each training phase, we train the model for one epoch on the public datasets (refer to supplementary materials for more details).
For multi-modal reasoning scenario, we take StructEqTable~\cite{docgenome}, ChartVLM~\cite{chartvlm} and Math-CLIP~\cite{mavis} as table, chart and math expert respectively.
For visual content extraction scenario, we employ GOT~\cite{got} as the document expert.
More detailed implementation specifics can be found in the supplementary material.

\vspace{-2pt}
\subsection{Comparison on Multi-modal Reasoning}
\label{subsec:multi-modal reasoning}
\vspace{-4pt}

\noindent \textbf{Comparison with Generalist Models.}
LMMs, such as LLaVA-OneVision~\cite{llava-onevision}, Qwen2-VL~\cite{qwen2-vl},InternVL2~\cite{internvl-1.5} and GPT-4o demonstrate powerful multi-modal reasoning abilities on general purpose scenarios.
However, these generalist models always exhibit limited performance when handling tasks under professional scenarios.
Our model demonstrates exceptional performance on the challenging multi-modal reasoning benchmarks MathVista and MathVerse, significantly outperforming existing generalist models. 
As shown in Tab.~\ref{tab:mathvista} and Tab.~\ref{tab:mathverse}, Chimera-8B achieved overall accuracies of 64.9 and 32.4, respectively, setting a new state-of-the-art (SOTA) for LMMs under the 70B scale.
Chimera-2B and Chimera-4B both significantly outperform the baseline InternVL2 series and achieve results comparable to models of much larger scales.
On MathVista, Chimera-8B stands out among both closed-source and open-source LMMs of the same size, leading GPT-4o by 1.1\%, and outperforming Qwen2-VL and InternVL by 6.7\% and 3.9\%, respectively. 
On MathVerse, Chimera-8B is only slightly behind GPT-4V, and surpasses the hundred-billion-scale LLaVA-NeXT by 7.9 points. This demonstrates that our approach, by integrating domain knowledge from different expert models, effectively enhances performance in specialized domains.

\noindent \textbf{Comparison with Specialist Models.}
Compared to specialist models such as Math-LLaVA~\cite{mathllava}, Math-PUMA~\cite{math-puma}, MAVIS~\cite{mavis}, and GeoX~\cite{xia2024geox}, Chimera demonstrates outstanding performance. 
As reported in Tab.~\ref{tab:mathvista} and Tab.~\ref{tab:mathverse}, Chimera-8B outperforms the previous best expert models by 18.3\% and 4.9\% on MathVista and MathVerse, respectively.
In contrast, the latest expert model, Math-PUMA~\cite{math-puma}, achieves performances of 44.7 and 31.8 on the two benchmarks, which is notably inferior to our method.
It is worth noting that these expert models have limited generalization ability and cannot handle tasks from other domains. In contrast, our model excels across various tasks in the table, chart, and document domains, proving Chimera’s powerful versatility.

\noindent \textbf{Further Improve Reasoning Ability with Preference Optimization.}
The emergence of preference optimization like RLHF has revolutionized model enhancement strategies. Chimera is able to capitalize on this trend through seamless preference optimization training integration, demonstrating remarkable scalability.
We construct 60K preference pairs using publicly available datasets and Chimera's outputs and perform a naive Direct Preference Optimization to develop Chimera$^\dagger$. Specific data construction process can be found in the supplementary material. 
Chimera$^\dagger$, with naive post-train, surpasses SOTA post-training methods by 1.0\%, shows significant 3.4\% improvement over base models without post-training, highlighting the framework’s scalability and potential.

\noindent \textbf{Fine-Grained Analysis.}
We manually classified MathVista questions by the domain of visual content and present the model's performance across domains in Tab.~\ref{tab:mathvista_domain}. 
In most cases, the model outperforms its baseline in each domain, further demonstrating that incorporating expert models enhances the generalist model’s performance on specialized tasks.
We also observed that expert models improve performance in general scenarios, suggesting that domain knowledge provides diverse insights for language model in handling visual information, thereby enhancing the model even on scenarios where experts are not activated.

Chimera-8B performs similarly in the chart domain but slightly worse in the table domain. 
This is due to the over-specialized function of the table expert, which, despite supporting comprehensive extraction, may introduces noise for the 8B baseline model because of the considerable task gap. 
In contrast, the chart expert's pre-training task covers both extraction and perception, with minimal impact. The math expert consistently improves performance across models due to its alignment with reasoning tasks.

\subsection{Comparison on Visual Structural Extraction}
\label{subsec:visual content extraction}
\noindent \textbf{Comparison with Generalist Models.}
For specialized tasks beyond VQA, generalist models similarly show limited performance. Tab.~\ref{tab:chartse} and Tab.~\ref{tab:tablese} present the results of visual structural extraction in the Chart and Table domains, respectively.
On ChartQA-SE and PlotQA-SE, Chimera outperforms representative generalist models such as GPT-4V, Qwen-VL~\cite{qwen-vl}, and InternVL-2~\cite{internvl-1.5} across the AP@strict, AP@slight, and AP@high metrics. 
In Table-SE, Chimera leads InternVL2 and Qwen2-VL by a larger margin in Edit Distance and TED scores, demonstrating strong domain-specific capability. 

Tab.~\ref{tab:inhouse-page} and Fig.~\ref{fig:in-house-page} exhibit results on Doc-SE, Chimera significantly outperforms InternVL2~\cite{internvl-1.5} across four metrics in bilingual tasks and shows balanced performance across different document categories.

\noindent \textbf{Comparison with Specialist Models.}
Compared to specialist models majoring in single task, Chimera still demonstrates strong performance. Specifically, for ChartQA-SE and PlotQA-SE, Chimera achieves excellent or competitive results across three metrics compared to the SOTA expert model GOT.
In Table-SE, Chimera also achieves comparable TED scores and outperforms with a lower Edit Distance by 0.092. 
As for Doc-SE, Chimera leads in most metrics for both English and Chinese documents, showing better overall generalization across document categories than GOT.


\begin{figure}[t]
    \vspace{-6pt}
    \centering
    \includegraphics[width=0.75\linewidth]{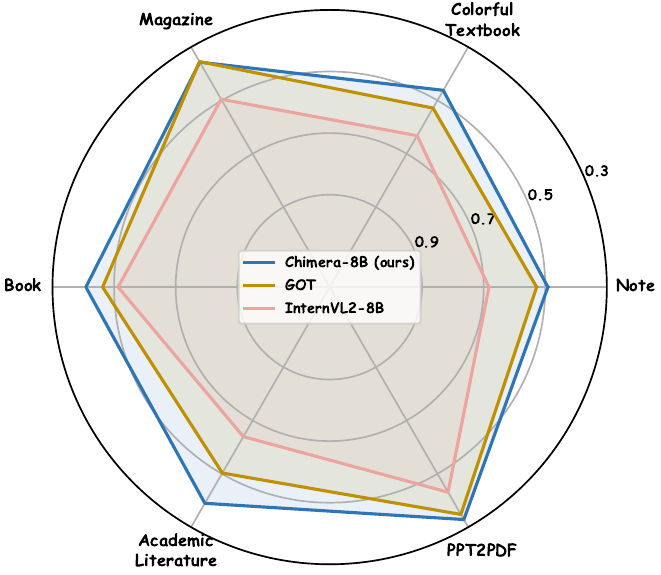}
    \vspace{-2pt}
    \caption{Comparison of Edit Distance $\downarrow$ across different document categories on Document Structural Extraction (Doc-SE) task.}
    \label{fig:in-house-page}
\vspace{-15pt}
\end{figure}

\subsection{Further analysis}
\label{subsec:further_analysis}

\begin{table}[h!]
\tiny 
\centering
\renewcommand\tabcolsep{3pt} 
\renewcommand\arraystretch{1.05} 
\resizebox{0.8\columnwidth}{!}{
\begin{tabular}{c|cccc}
\toprule
\textbf{GT \textbackslash Prediction} & \textbf{General} & \textbf{Table} & \textbf{Chart} & \textbf{Math} \\
\midrule
\textbf{General} & \textbf{--} & 0 & 16 & 6 \\
\textbf{Table}   & 1 & \textbf{--} & 0 & 0 \\
\textbf{Chart}   & 1 & 0 & \textbf{--} & 0 \\
\textbf{Math}    & 22 & 0 & 0 & \textbf{--} \\
\bottomrule
\end{tabular}
}
\vspace{-4pt}
\caption{Error statistics of router on MathVista \textit{testmini}.}
\label{tab:router_mistake}
\end{table}

\noindent \textbf{Effects of GSCM.}
To assess whether our proposed GSCM enhances alignment with the expert model, we analyze the attention distribution of the model’s output across general visual tokens $\mathcal{H}_v$ and expert visual tokens $\mathcal{H}_e$.
As illustrated in \cref{fig:heatmap}, Chimera places greater emphasis on domain features during inference, demonstrating improved alignment between the specialist models and the generalist LMM. 
In contrast, Chimera w/o masking exhibits significantly weaker utilization of domain features. This occurs because the general encoder, which is already well-aligned with the language model, often results in imbalanced optimization between the general encoder and expert models. 

\begin{figure}
\vspace{4pt}
    \centering
    \includegraphics[width=\linewidth]{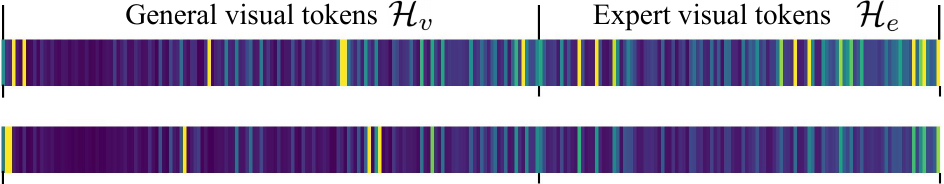}
    \vspace{-18pt}
    \caption{Attention distribution on Chimera with (Top) \& without (Bottom) masking.}
    \vspace{-16pt}
    \label{fig:heatmap}
\end{figure}

\noindent \textbf{Analysis about Expert Router Error.}
Due to the significant differences in visual inputs across different domains, Chimera can achieve effective classification using a simple linear layer, thereby guiding expert model selection.
As shown in \cref{tab:router_mistake}, it achieves 95.4\% accuracy on MathVista, proving the router's effectiveness. 
All errors stem from confusion between the general and expert domains, not between experts. This issue arises because training labels are dataset-based, and the ``general" category includes mixed-domain images. Explicit domain annotation for each image could better address this.

\vspace{-2pt}
\section{Conclusion}
\label{sec:conc}
\vspace{-2pt}

We present Chimera, a scalable pipeline that integrates specialist models into generalist LMMs, enabling adaptation to specialized tasks. Our approach transforms LMMs, like InternVL-2, into versatile models capable of handling tasks across tables, math, documents, \textit{etc}. Chimera pioneers new directions for bridging generalist and specialist models.


\section*{Acknowledgements}
The research was supported by Shanghai Artificial Intelligence Laboratory, a locally commissioned task from the Shanghai Municipal Government, the Shanghai Municipal Science and Technology Major Project, and Shanghai Rising Star Program (Grant No. 23QD1401000).

This work is supported by National Key Research and Development Program of China (No. 2022ZD0160101), Shanghai Natural Science Foundation (No. 23ZR1402900),  Shanghai Municipal Science and Technology Major Project (No.2021SHZDZX0103). The computations in this research were performed using the CFFF platform of Fudan University.

\clearpage
\setcounter{page}{1}
\maketitlesupplementary

Due to the eight-page limitation of the main text, we provide more details and visualizations from the following aspects:

        

\begin{itemize}
    \item Sec.~\ref{app:task_expert}: Selection strategy for pre-training tasks and expert models.
    \item Sec.~\ref{app:dataset_details}: Dataset Details in training.
    \item Sec.~\ref{app:general_task}: Chimera's performance on general tasks.
    \item Sec.~\ref{app:dpo_details}: Details about preference optimization.
    \item Sec.~\ref{app:mask_ratio}: Experiment results on mask ratio selection.
    \item Sec.~\ref{app:tab-doc}: Introduction of Table-SE and Doc-SE.
    \item Sec.~\ref{app:exps}: Experiments of scaling up more experts.
    \item Sec.~\ref{app:comp}: Comparison with existing works.
    \item Sec.~\ref{app:imp}: More information of implementation details.
    \item Sec.~\ref{app:vis}: Visualization of Chimera's visual content extraction performance.
\end{itemize}

\section{Pre-training Tasks and Expert Models}
\label{app:task_expert}

\begin{figure}[h]
\vspace{-4pt}
    \centering
    \includegraphics[width=0.92\linewidth]{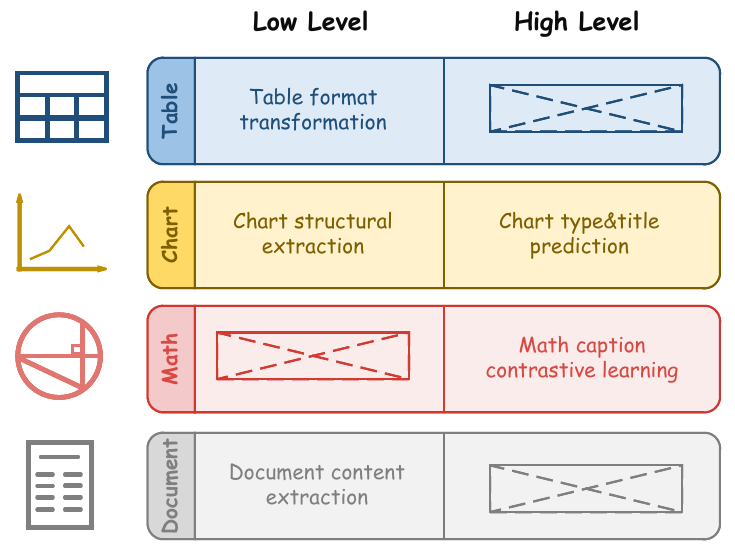}
    \vspace{-10pt}
    \caption{Pre-training tasks of expert models considered by Chimera.}
    \label{fig:expert selection}
\vspace{-10pt}
\end{figure}

The type of pre-training task significantly affects model performance, which we consider when selecting expert models. As shown in Fig.~\ref{fig:expert selection}, we categorize low-level tasks as the precise extraction of domain-specific visual content and structure (\textit{e.g.}, Table2LaTeX, Chart2Markdown, Doc2Markdown), while high-level tasks involve understanding and summarizing image content.
We select expert models with diverse pre-training task configurations. For the table expert, we use the encoder from StructEqTable~\cite{docgenome}, which effectively converts table images into LaTeX/HTML. For the chart expert, we choose the encoder from ChartVLM~\cite{chartvlm}, which excels in structural extraction and chart type classification. For the math expert, we adopt Math-CLIP~\cite{mavis}, trained on extensive geometry and function caption data. For document structural extraction, we employ the encoder from the latest model, GOT~\cite{got}.

\section{Dataset Details}
\label{app:dataset_details}

The datasets used for Chimera is presented in \cref{tab:reasoner_data} and \cref{tab:extractor_data}. All the datasets we used come from publicly accessible datasets.
\begin{table}[t]
\centering
\renewcommand\tabcolsep{3pt} 
\renewcommand\arraystretch{1.05} 
\resizebox{\columnwidth}{!}{
\begin{tabular}{l|l}
\toprule
\multirow{8}{*}{\textit{\textbf{Stage 1}}} & \textit{\textbf{General:}} \\ 
& ShareGPT4v~\cite{chen2023sharegpt4v}, ShareGPT4-o~\cite{chen2023sharegpt4v}    \\
\cmidrule{2-2}

& \textit{\textbf{Table:}} \\ 
& TableX~\cite{docgenome}  \\
\cmidrule{2-2}

& \textit{\textbf{Chart:}} \\ 
& ChartQA~\citep{masry2022chartqa}, PlotQA~\citep{methani2020plotqa},ChartX~\citep{chartvlm}, SimChart~\citep{xia2023structchart}   \\
\cmidrule{2-2}

& \textit{\textbf{Math:}} \\ 
& MAVIS-Caption~\cite{mavis} \\
\midrule
\multirow{19}{*}{\textit{\textbf{Stage 2:}}} & \textit{\textbf{Language:}} \\ 
& Kaggle-science-exam~\cite{kaggle-llm-science-exam}, MathInstruct~\cite{yue2023mammoth}, \\
& MathQA~\cite{amini2019mathqa}, SciInstruct~\cite{zhang2024sciglm}, Orcamath~\cite{mitra2024orcamath} \\
\cmidrule{2-2}

& \textit{\textbf{General:}} \\ 
& ShareGPT4v~\cite{chen2023sharegpt4v}, ShareGPT4-o~\cite{chen2023sharegpt4v}, LLaVAR~\cite{zhang2023llavar}, \\
& AI2D (GPT4V)~\cite{llava-onevision}, AI2D (InternVL~\cite{internvl-1.5}), AI2D (Original)~\cite{kembhavi2016diagram}, \\
& MathVision~\cite{wang2024measuring}, IconQA~\cite{lu2021iconqa}, MapQA~\cite{chang2022mapqa}, ScienceQA~\cite{saikh2022scienceqa},  \\
& ArxivQA~\cite{li2024multimodal}, TQA~\cite{kembhavi2017you}, CLEVR-Math~\cite{johnson2017clevr}, Super-CLEVR~\cite{li2023super}, \\
& Cambrian Data Engine~\cite{cambrian} \\
\cmidrule{2-2}

& \textit{\textbf{Table:}} \\ 
& TableX~\cite{docgenome}, TabMWP~\cite{lu2023dynamic}, MMTab~\cite{tablellava}   \\
\cmidrule{2-2}

& \textit{\textbf{Chart:}} \\ 
& PlotQA~\citep{methani2020plotqa},ChartX~\citep{chartvlm}, SimChart~\cite{xia2023structchart}, Chart2Text~\cite{kantharaj2022chart}, \\
& ChartQA~\citep{masry2022chartqa},  LRV Chart~\cite{liu2023aligning}, ChartGemma~\cite{chartgemma}, DVQA~\cite{kafle2018dvqa}, \\
& FigureQA~\cite{kahou2017figureqa},  VisText~\cite{tang2023vistext} \\
\cmidrule{2-2}

& \textit{\textbf{Math:}} \\ 
& MAVIS-Caption~\cite{mavis}, Geo170K~\cite{gllava}, GeoMVerse~\cite{kazemi2023geomverse}, \\
& MAVIS Manual Collection~\cite{mavis}, MAVIS Data Engine~\cite{mavis} \\
& Geometry3K~\cite{lu2021inter}, GeoQA+~\cite{chen2021geoqa}, InterGPS~\cite{lu2021inter} \\
\bottomrule
\end{tabular}
}
\vspace{-8pt}
\caption{Dataset used for \ptsModify{multi-modal reasoning scenario}. \textit{\textbf{Stage 1}} and \textit{\textbf{Stage 2}} represent Domain-General Knowledge Alignment and Visual Instruction Tuning separately.}
\label{tab:reasoner_data}
\end{table}

\begin{table}[t]
\centering
\renewcommand\tabcolsep{3pt} 
\renewcommand\arraystretch{1.05} 
\resizebox{\columnwidth}{!}{
\begin{tabular}{l|l}
\toprule
\multirow{1}{*}{\textit{\textbf{Stage 1}}} & ChartQA~\citep{masry2022chartqa}, PlotQA~\citep{methani2020plotqa},ChartX~\citep{chartvlm},, SimChart~\citep{xia2023structchart}, TableX~\cite{docgenome}\\
\midrule
\multirow{1}{*}{\textit{\textbf{Stage 2}}} & DocGenome~\cite{docgenome}, DocStruct4M~\cite{hu2024mplug}, DocVQA~\cite{mathew2021docvqa} \\
\bottomrule
\end{tabular}
}
\vspace{-8pt}
\caption{Datasets used for \ptsModify{visual content extraction scenario}. \textit{\textbf{Stage 1}} represents Domain-General Knowledge Alignment, and \textit{\textbf{Stage 2}} represents Visual Instruction Tuning.}
\label{tab:extractor_data}
\vspace{-12pt}
\end{table}

\section{Evaluation on General Tasks}
\label{app:general_task}
\begin{table}[h]
    \centering
    \small
    \vspace{-2pt}
    \begin{tabular}{lcccc}
        \toprule
         & \multicolumn{2}{c}{\textbf{InternVL2}} & \multicolumn{2}{c}{\textbf{Chimera}} \\
        \cmidrule(lr){2-3} \cmidrule(lr){4-5}
        & 4B & 8B & 4B & 8B \\
        \midrule
        \textbf{Existence} & 200.00 & 190.00 & 200.00 & 195.00 \\
        \textbf{Count} & 123.33 & 158.33 & 130.00 & 155.00 \\
        \textbf{Position} & 143.33 & 155.00 & 123.33 & 148.33 \\
        \textbf{Color} & 165.00 & 175.00 & 160.00 & 190.00 \\
        \textbf{Posters} & 158.84 & 168.03 & 159.86 & 164.97 \\
        \textbf{Celebrity} & 125.00 & 148.53 & 145.29 & 162.65 \\
        \textbf{Scene} & 158.75 & 152.50 & 163.50 & 157.75 \\
        \textbf{Landmark} & 167.25 & 178.25 & 167.25 & 177.75 \\
        \textbf{Artwork} & 144.75 & 154.50 & 144.00 & 153.00 \\
        \textbf{OCR} & 147.50 & 162.50 & 117.50 & 132.50 \\
        \bottomrule
    \end{tabular}
    \vspace{-10pt}
\caption{Performance on perception sub-tasks of MME.}
\vspace{-13pt}
\label{tab:MME results}
\end{table}

We evaluate Chimera's general capabilities using the perception set from the general benchmark MME~\cite{fu2023mme}, with results presented in \cref{tab:MME results}.
Across different model sizes, Chimera and InternVL exhibit varying strengths across different tasks, achieving overall comparable performance. This suggests that the Chimera framework introduces minimal degradation to the model’s general task capabilities.
Meanwhile, Chimera demonstrates strong expertise in domains such as tables, math, charts, and documents, further validating that our proposed approach effectively enhances a generalist LMM’s domain-specific knowledge without compromising its general performance.

\section{Details about preference optimization}
\label{app:dpo_details}

For preference optimization, we adopt a commonly used approach: we randomly sample 10k problems from MathV-360K, generating 16 responses per problem using Chimera. 
Each response is classified based on correctness using rule-based answer matching, and after filtering, we construct 60k preference pairs for Direct Preference Optimization (DPO) training. Then we perform DPO training on 60K data for 1 epoch.

\section{Mask Ratio Selection}
\label{app:mask_ratio}
\begin{table}[h!]
\vspace{-6pt}
\centering
\renewcommand\tabcolsep{3pt} 
\renewcommand\arraystretch{1.05} 
\tiny 
\resizebox{\columnwidth}{!}{
\begin{tabular}{l|c|c|cccc}
\toprule
Model & Ratio & ALL & General &Chart &Table &Math \\ 
\midrule
InternVL2-4B~\citep{internvl-1.5} & N/A & 57.0 & 50.1 & 66.2 & 65.7 & 58.3 \\ 
InternVL2-4B-NF~\citep{internvl-1.5} & N/A & 58.5 & 51.5 & 67.1 & \textbf{74.3} & 58.6 \\
Chimera-4B-0.0 & 0.0 & 59.4 & 50.8 & 66.2 & 67.1 & 65.5 \\
Chimera-4B & 0.3 & \textbf{61.3} & \textbf{54.0} & 64.8 & 72.9 & \textbf{66.9} \\ 
Chimera-4B-0.5 & 0.5 & 60.4 & 51.3 & \textbf{68.5} & 70.0 & 65.8 \\
Chimera-4B-1.0 & 1.0 & 56.2 & 51.5 & 63.5 & 72.9 & 53.6 \\
\bottomrule[.7pt]
\end{tabular}
}
\vspace{-10pt}
\caption{Ablation results on different visual content domain on the \textit{testmini} subset of MathVista. InternVL2-4B-NF represents naive finetune of baseline with same settings, Chimera-4B-$R$ means Chimera model trained with mask ratio $R$ in GSCM.}
\label{tab:ablation}
\vspace{-6pt}
\end{table}
We conducted an ablation study on 4B scale models to assess our approach's effectiveness, as shown in Table~\ref{tab:ablation}. It should be noted that model with mask ratio $1.0$ does not have access to the general encoder during training, contrary to our intentions. Thus, we modified this case to give the model an 80\% probability of masking all general features.
The results show that naively finetuning the LMM leads to limited performance improvement. By incorporating domain knowledge from expert models, even the case without GSCM still yields better results than naive finetuning.
As the mask ratio increases, the model's performance improves initially and then declines. This indicates that slightly masking helps balance encoder optimization, leading to better alignment. However, as the mask ratio increase, we believe excessive masking prevents the model from effectively learning to utilize both features for reasoning.
Based on the above observations, we set the mask ratio to 0.3 in Chimera's implementation.
We also observed that performance trends vary across domains as the mask ratio changes, suggesting that the alignment difficulty of expert models differs by domain and task, which we leave for future exploration.

\section{Details of Table-SE and Doc-SE}
\label{app:tab-doc}

In Tables 7 and 8 of the main text, we conduct the experiments on Table Structural Extraction (Table-SE) task and Document Structural Extraction (Doc-SE) task, respectively. In this section, we primarily introduce the evaluation dataset construction method and provide detailed information about the dataset.

\subsection{Data Source}
\begin{table}[h!]
\centering
\small
\renewcommand{\arraystretch}{0.85}
\begin{tabular}{l|c}
\toprule
 & \textbf{Count} \\
\midrule
\multicolumn{2}{l}{\textbf{Document Categories}} \\
PPT2PDF & 43 \\
Academic Literature & 42 \\
Book & 13 \\
Colorful Textbook & 37 \\
Magazine & 30 \\
Exam Paper & 7 \\
Note & 18 \\
Newspaper & 15 \\
\midrule
\multicolumn{2}{l}{\textbf{Language}} \\
Simplified Chinese & 128 \\
English & 77 \\
\midrule
\multicolumn{2}{l}{\textbf{Layout}} \\
1 and More Column & 27 \\
Single Column & 91 \\
Other Layout & 43 \\
Double Column & 40 \\
Three Column & 4 \\
\midrule
\# Total & 205 \\

\bottomrule
\end{tabular}
\vspace{-4pt}
\caption{Statistical information of Doc-SE.}
\label{tab:docse_detail}
\end{table}

\begin{table}[h!]
\centering
\renewcommand{\arraystretch}{0.8} 
\small 
\begin{tabular}{l|r}
\toprule
 & \textbf{Count} \\
\midrule
\multicolumn{2}{l}{\textbf{Background}} \\
w/o Background & 80 \\
w/ Background & 20 \\
\midrule
\multicolumn{2}{l}{\textbf{Equation}} \\
w/o Equation & 78 \\
w/ Equation & 22 \\
\midrule
\multicolumn{2}{l}{\textbf{Language}} \\
English & 45 \\
English \& Chinese Mixed & 5 \\
Chinese & 50 \\
\midrule
\multicolumn{2}{l}{\textbf{Table Format}} \\
Three-line Table & 47 \\
Full-bordered Table & 39 \\
Partial-bordered Table & 14 \\
w/o Merged Cells & 58 \\
w/ Merged Cells & 42 \\
\midrule
\multicolumn{2}{l}{\textbf{Layout}} \\
Horizontal & 97 \\
Vertical & 3 \\
\midrule
\# Total & 100\\
\bottomrule
\end{tabular}
\vspace{-4pt}
\caption{Statistical information of Table-SE.}
\label{tab:tablese_detail}

\end{table}

Our benchmark was developed through a systematic sampling process from an initial collection of $200,000$ PDF documents sourced from Common Crawl, Google, Baidu search engines, and internal repositories. We initially extracted visual features using ResNet-50~\citep{he2016deep} and performed clustering algorithm using Faiss~\cite{johnson2019billion} to identify diverse document patterns. From the 10 cluster centers, we sampled 6,000 visually diverse pages, which were then manually annotated with attributes such as page type, layout type, and language. As illustrated in Table~\ref{tab:docse_detail} and Table~\ref{tab:tablese_detail}, the final benchmark includes 205 page-level PDF images and 100 table images, ensuring comprehensive representation of real-world document scenarios with various layouts and attributes.

\subsection{Annotation Process}
For ensuring annotation quality and efficiency, we design separate standardized processes for page-level PDF documents and tables.

For page-level PDF documents, our process consists of three stages: (1) We first employ fine-tuned LayoutLMv3 for layout detection and PaddleOCR for text recognition as intelligent pre-annotation. (2) Professional annotators then refine the detection boxes, verify text content accuracy, and enhance annotations with reading order and affiliation details. (3) Finally, researchers review the annotations to ensure overall quality and accuracy.

For table annotations, we follow a similar but specialized three-stage approach: (1) We utilize GPT-4o and PaddleOCR for initial table annotations. (2) Annotators then verify and correct the table structure and content, using specialized tools like Tables Generator for verification. (3) Finally, experts through table annotations re-rendering to ensure correct HTML and LaTeX code labels.

\subsection{Showcase}
\begin{figure*}
    \centering
    \includegraphics[width=0.9\linewidth]{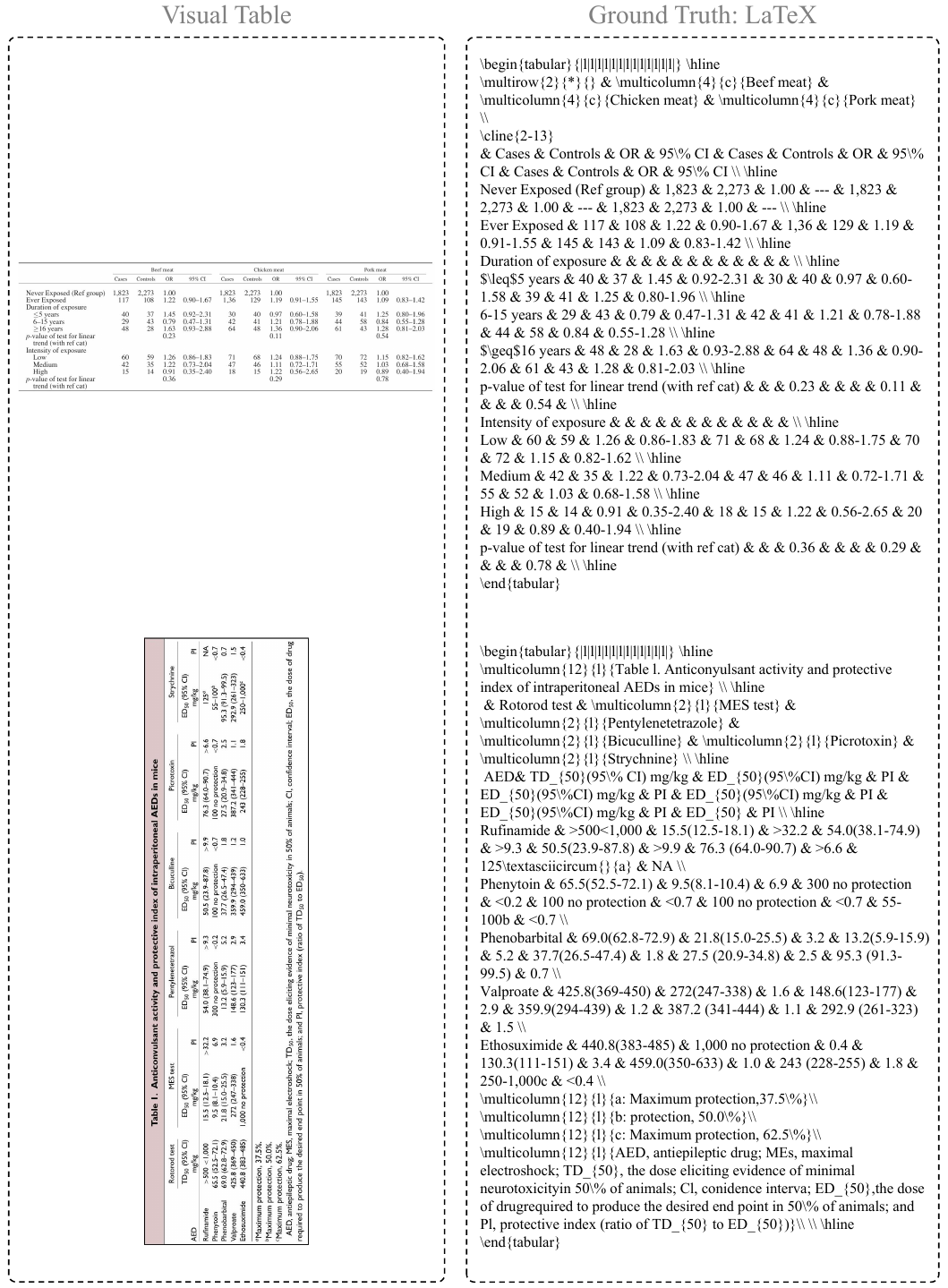}
    \caption{Showcase of Table-SE.}
    \label{fig:tablese_showcase_1}
\end{figure*}
\begin{figure*}
    \centering
    \includegraphics[width=\linewidth]{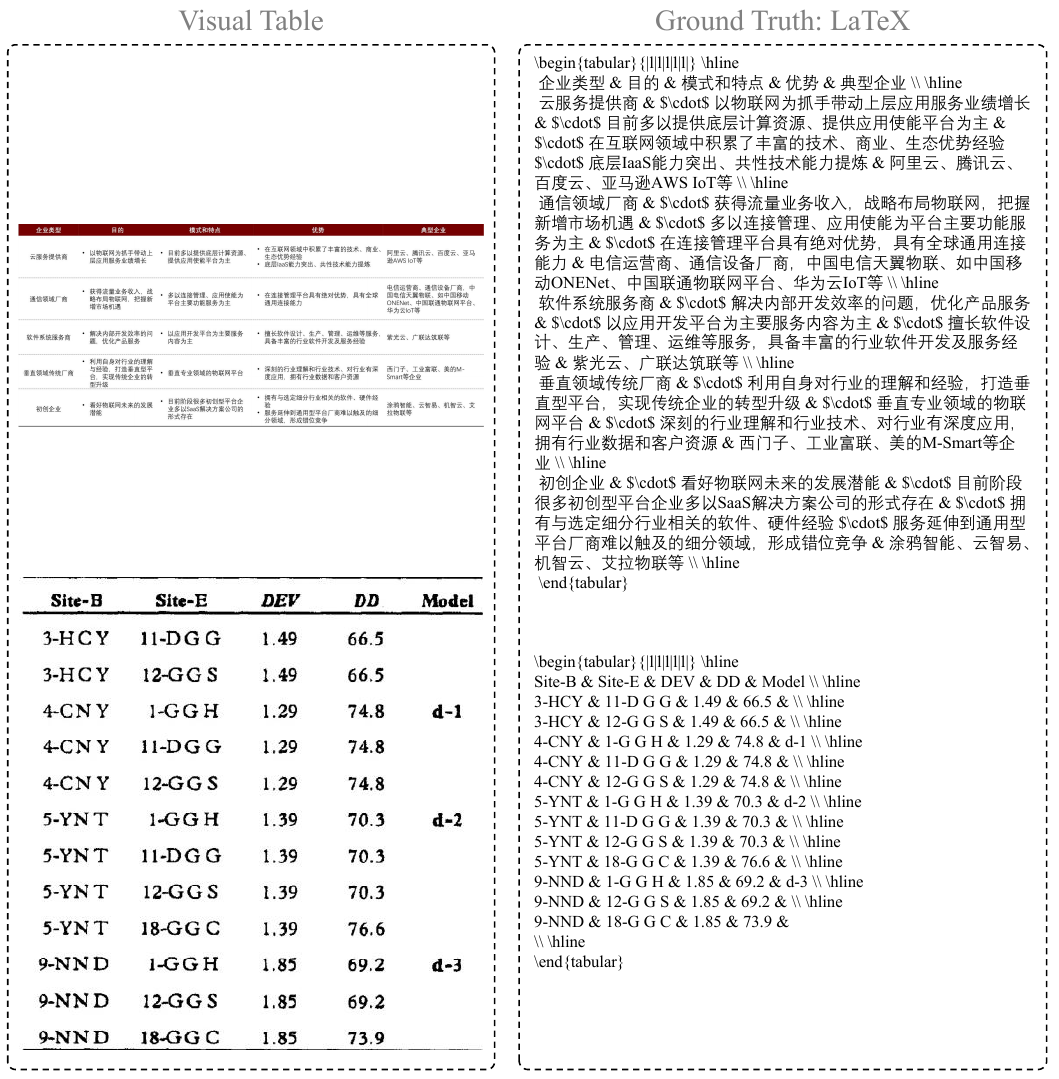}
    \caption{Showcase of Table-SE.}
    \label{fig:tablese_showcase_2}
\end{figure*}

We provide several visualization examples of Table-SE in Fig.~\ref{fig:tablese_showcase_1} and Fig.~\ref{fig:tablese_showcase_2}, where each item contains a visual table and its corresponding LaTeX code.

\section{Experiments of Scaling Up More Experts}
\label{app:exps}
\begin{table}[ht!]
\centering
\renewcommand\tabcolsep{3pt} 
\renewcommand\arraystretch{1.0} 
\tiny 
\resizebox{1.0\columnwidth}{!}{
\begin{tabular}{c|c|cccc}
\toprule
Model & ALL & General &Chart &Table &Math \\ 
\midrule
InternVL2-4B & 57.0 & 50.1 & 66.2 & 65.7 & 58.3 \\ 
InternVL2-4B w/ Chart Expert & 59.4 & 52.0 & 68.0 & 72.9 & 60.8 \\
Chimera-4B & 61.3 & 54.0 & 64.8 & 72.9 & 66.9 \\ 
\bottomrule
\end{tabular}
}
\vspace{-4pt}
\caption{Accuracy scores of different visual content domain on the \textit{testmini} subset of MathVista.Those do not belong to the last three domains are uniformly classified as General for simplicity. InternVL2-4B w/ Chart Expert represent the case only integrating chart expert model.}
\label{tab:mathvista_domain_scaleup}
\end{table}
To further validate the impact of scaling up the number of expert models, we provide ablation results introducing only the chart expert. In this case, non-chart data is encoded solely by the general encoder during training.
As shown in Table~\ref{tab:mathvista_domain_scaleup}, incorporating only the chart expert obtains lower MathVista~\cite{lu2023mathvista} overall score by 1.9 points than Chimera-4B. 

Specifically, InternVL2-4B w/ Chart Expert also shows improvements in general scenarios, though less significant than Chimera, which integrates three expert models.
In the chart domain, InternVL2-4B w/ Chart Expert achieves notable gains by avoiding conflicts among multiple experts with large task gaps. However, Chimera's integration of multiple experts enhances performance across diverse domains, boosting overall results.
In the math domain, InternVL2-4B w/ Chart Expert scores 6.1 points lower than Chimera, demonstrating the strong mathematical reasoning capabilities derived from integrating the math expert.

\section{Comparison with existing works}
\label{app:comp}
Integrating specialist experts that contain specialized prior knowledge presents a promising approach to improve the specific capabilities of generalist model.
MoVA~\cite{zong2024mova} proposes a multi-turn method that relies on the language model to call an expert in the first round and generates responses in the second, which reduces conciseness and efficiency.
MoME~\cite{shen2024mome} uses soft-weighting to fuse multiple visual encoders, enabling VLMs to benefit from leveraging representations from different encoders. However, this approach lacks explicit guidance for encoder selection and introduces additional concerns such as inference efficiency and uniform visual feature sizes.

\section{Training Configuration}
\label{app:imp}

\begin{table*}[htp]
    \centering
    \hfill
    \setlength{\tabcolsep}{12pt}
    \renewcommand{\arraystretch}{1.2}
    \resizebox{\textwidth}{!}{%
    \begin{tabular}{@{}l|ll|c|c}
    \toprule
    & & & \textbf{Domain-General Knowledge Alignment} & \textbf{Visual Instruction Tuning} \\
    \midrule 
    
    \multirow{9}{*}{\rotatebox[origin=c]{90}{\footnotesize \textit{Vision}}}
    
    & \multirow{4}{*}{\textbf{Resolution}} & General Encoder $E^g$ & 448 & 448$\times$\{\{1,2,3,4,5,6\}$\times$1, 1$\times$\{2,3,4,5,6\}, 2$\times$ \{2,3\}, 3$\times$ 2 \}\}  \\
    & & Table Encoder $E^{table}$ & N/A & N/A \\
    & & Chart Encoder $E^{chart}$ & N/A & N/A \\
    & & Math Encoder $E^{math}$ & 336 & 336  \\
    \cmidrule{2-5}
    
    & \multirow{4}{*}{\textbf{\#Tokens}} & General Encoder $E^g$ & 256 & Max 256$\times$6  \\
    & & Table Encoder $E^{table}$ & 2048 & 2048 \\
    & & Chart Encoder $E^{chart}$ & 2048 & 2048 \\
    & & Math Encoder $E^{math}$ & 576 & 576  \\
    \cmidrule{2-5}
    & \multicolumn{2}{c|}{\textbf{Total Tokens}}  & 256 + \{0, 2048, 2048, 576\} & Max 256$\times$6 + \{0, 2048, 2048, 576\}  \\

    \midrule 
    
    
    \multirow{10}{*}{\rotatebox[origin=c]{90}{\footnotesize \textit{Training}}}
    & \multicolumn{2}{c|}{\textbf{\#Samples}} & 1.1M & 2.6M \\
    & \multicolumn{2}{c|}{\textbf{GSCM ratio}} & N/A & 0.3 \\
    & \multicolumn{2}{c|}{\textbf{Dynamic High Res~\cite{internvl-1.5}}} & False & True \\
    & \multicolumn{2}{c|}{\textbf{Trainable}} & General Projector $P^g$, Expert Projector Set $S^e$ & General Projector $P^g$, Expert Projector Set $S^e$, Language Model $f$  \\
    & \multicolumn{2}{c|}{\textbf{Batch Size}} & 256/128 & 128 \\
    & \multicolumn{2}{c|}{\textbf{LR}} & 4e-5/2e-5 & 2e-5/1e-5 \\
    & \multicolumn{2}{c|}{\textbf{Warm Up}} & 100 steps &0.03 ratio \\
    & \multicolumn{2}{c|}{\textbf{LR Scheduler}} & Consine & Consine \\
    & \multicolumn{2}{c|}{\textbf{Max Length}} & 4096 & 8192 \\
    & \multicolumn{2}{c|}{\textbf{Weight Decay}} & 0.01 & 0.01 \\
    & \multicolumn{2}{c|}{\textbf{Epoch}} & 1 & 1  \\
    \bottomrule
    \end{tabular}
    }
    \caption{
    Detailed configuration for each training stage of \ptsModify{Chimera in multi-modal reasoning scenario}. The table outlines the progression of vision parameters, dataset characteristics and training hyperparameters. 
    For elements containing ``/", the left side represents configurations used by the 2B and 4B model, while the right side represents configurations used by the 8B model.
    }
    \label{tab:reasoner_config}
    \end{table*}

\begin{table*}[htp]
    \centering
    \hfill
    \setlength{\tabcolsep}{12pt}
    \renewcommand{\arraystretch}{1.2}
    \resizebox{\textwidth}{!}{%
    \begin{tabular}{@{}l|ll|c|c}
    \toprule
    & & & \textbf{Domain-General Knowledge Alignment} & \textbf{Visual Instruction Tuning} \\
    \midrule 
    
    \multirow{5}{*}{\rotatebox[origin=c]{90}{\footnotesize \textit{Vision}}}
    
    & \multirow{2}{*}{\textbf{Resolution}} & General Encoder $E^g$ & 448 & 448$\times$\{\{1,2,3,4,5,6\}$\times$1, 1$\times$\{2,3,4,5,6\}, 2$\times$ \{2,3\}, 3$\times$ 2 \}\}  \\
    & & Document Encoder $E^document$ & 1024 & 1024  \\
    \cmidrule{2-5}
    
    & \multirow{2}{*}{\textbf{\#Tokens}} & General Encoder $E^g$ & 256 & Max 256$\times$6  \\
    & & Document Encoder $E^document$ & 256 & 256  \\
    \cmidrule{2-5}
    & \multicolumn{2}{c|}{\textbf{Total Tokens}}  & 256 + 256 & Max 256$\times$6 + 256  \\

    \midrule 
    
    
    \multirow{10}{*}{\rotatebox[origin=c]{90}{\footnotesize \textit{Training}}}
    & \multicolumn{2}{c|}{\textbf{\#Samples}} & 995K & 275K \\
    & \multicolumn{2}{c|}{\textbf{GSCM ratio}} & N/A & 0.3 \\
    & \multicolumn{2}{c|}{\textbf{Dynamic High Res~\cite{internvl-1.5}}} & False & True \\
    & \multicolumn{2}{c|}{\textbf{Trainable}} & General Projector $P^g$, Expert Projector Set $S^e$ & General Projector $P^g$, Expert Projector Set $S^e$, Language Model $f$  \\
    & \multicolumn{2}{c|}{\textbf{Batch Size}} & 256 & 128 \\
    & \multicolumn{2}{c|}{\textbf{LR}} & 4e-5 & 2e-5 \\
    & \multicolumn{2}{c|}{\textbf{Warm Up}} & 100 steps &0.03 ratio \\
    & \multicolumn{2}{c|}{\textbf{LR Scheduler}} & Consine & Consine \\
    & \multicolumn{2}{c|}{\textbf{Max Length}} & 4096 & 8192 \\
    & \multicolumn{2}{c|}{\textbf{Weight Decay}} & 0.01 & 0.01 \\
    & \multicolumn{2}{c|}{\textbf{Epoch}} & 1 & 1  \\
    \bottomrule
    \end{tabular}
    }
    \caption{
    Detailed configuration for each training stage of \ptsModify{Chimera in visual content extraction scenario}. The table outlines the progression of vision parameters, dataset characteristics and training hyperparameters. 
    }
    \label{tab:extractor_config}
    
    \end{table*}


The training strategy is summarized in Tab~\ref{tab:reasoner_config} and Tab~\ref{tab:extractor_config}.
During the two-stage training process, we gradually increase the maximum image resolution and the number of visual tokens of the general visual encoder $E^g$.
In the Domain-General Knowledge Alignment stage, we use thumbnail images as input for $E^g$ without employing the widely-used Dynamic High Resolution (DHR) technique~\cite{llava-onevision,internvl-1.5}. In the Visual Instruction Tuning stage, DHR is introduced, allowing up to six times more visual tokens. 
At this stage, we apply the Generalist-Specialist Collaboration Masking (GSCM) mechanism with a masking ratio of 0.3 to constrain $E^g$, encouraging the model to leverage domain-specific information from expert models.
For trainable modules, the Domain-General Knowledge Alignment stage updates only the General Projector $P^g$ and Expert Projector Set $S^e$. In subsequent stages, the General Projector $P^g$, Expert Projector Set $S^e$, and Language Model $f$ are updated.

\section{Visualization of Chimera on Visual Content Extraction}
\label{app:vis}

\subsection{Table Format Transformation}
We provide the rendered table of the output results of Chimera-8B to show its table format transformation performance. 
As shown in Fig.~\ref{fig:table_demo_1}, Fig.~\ref{fig:table_demo_2} and Fig.~\ref{fig:table_demo_3}, 
Chimera excels in extracting and formatting table content from both Arxiv-style and more diverse table layouts with high accuracy.

\begin{figure*}
    \centering
    \includegraphics[width=\linewidth]{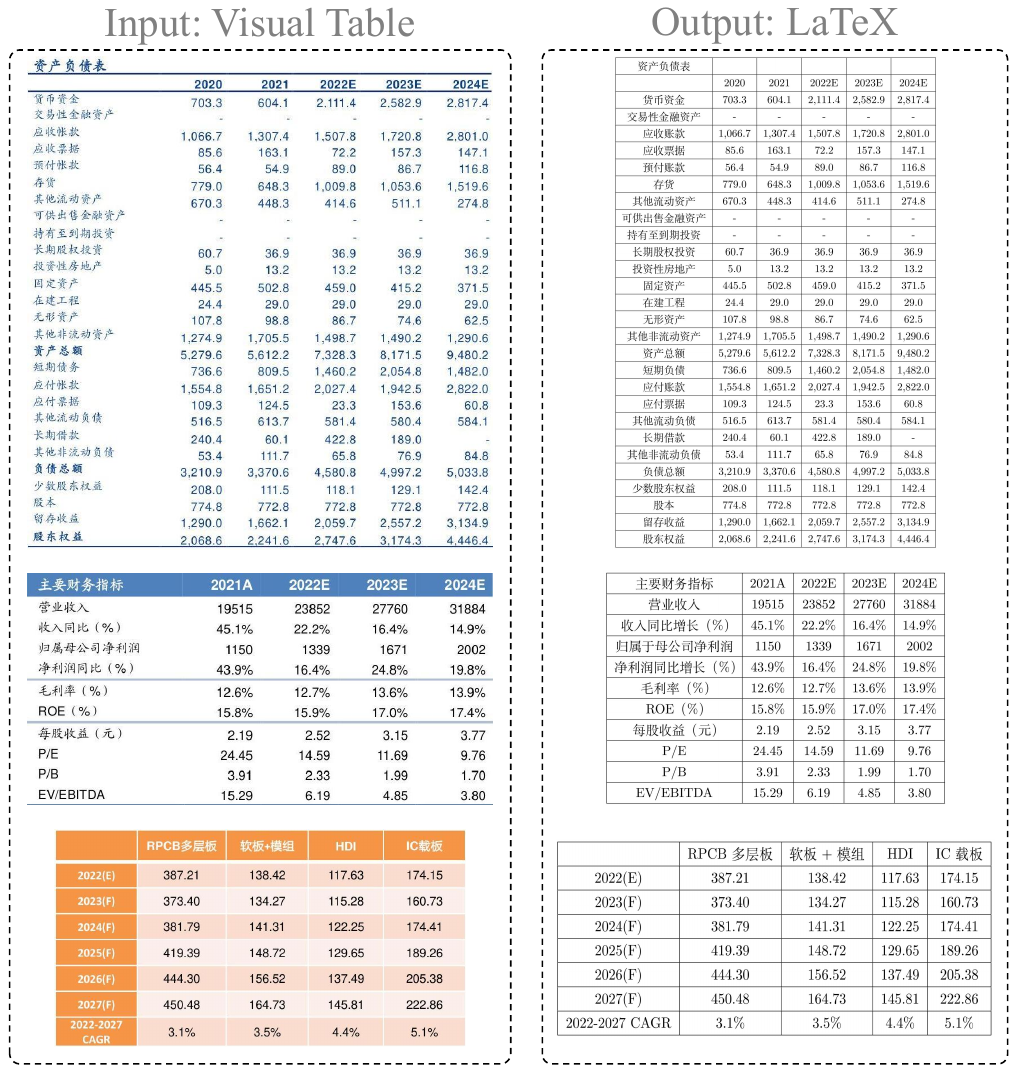}
    \caption{Output of Chimera-8B on Table Format Transformation.}
    \label{fig:table_demo_1}
\end{figure*}
\begin{figure*}
    \centering
    \includegraphics[width=\linewidth]{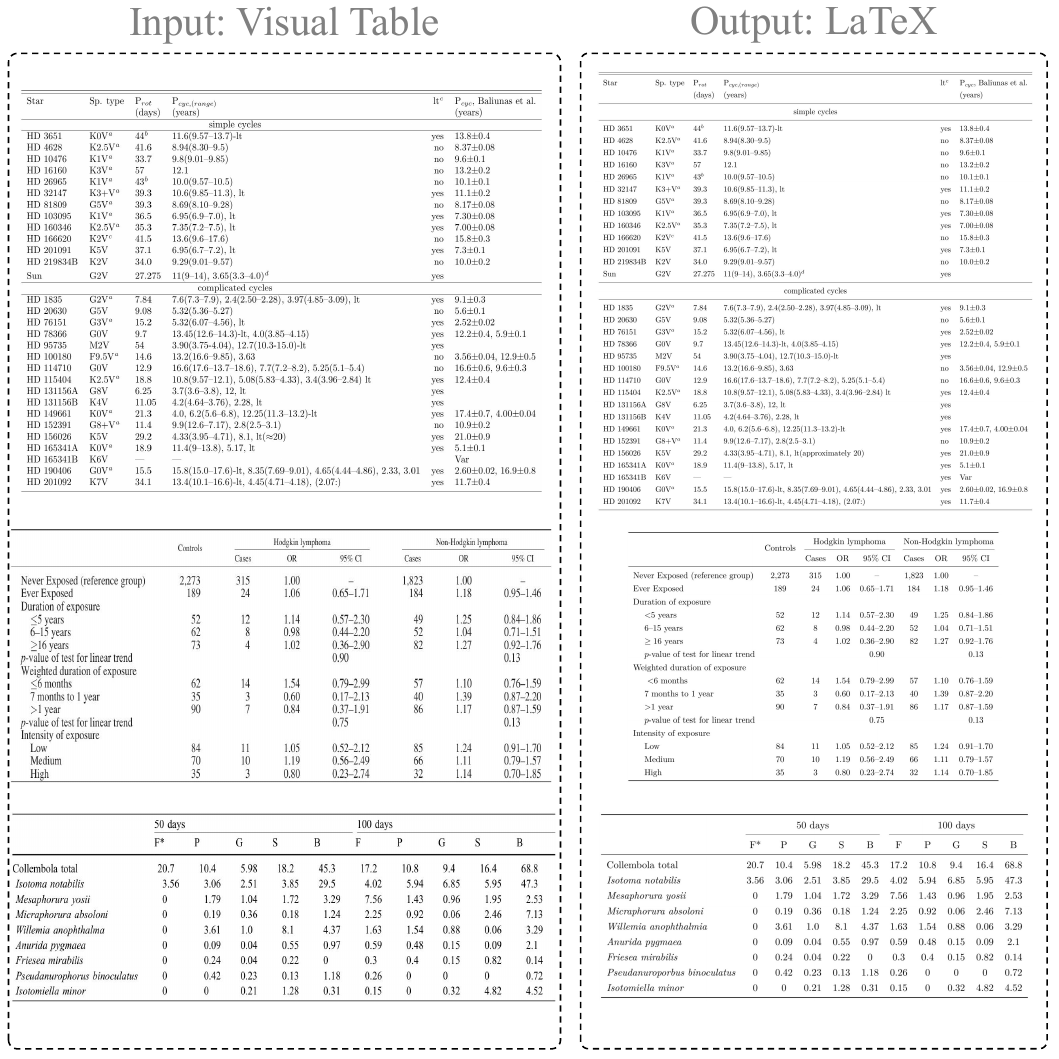}
    \caption{Output of Chimera-8B on Table Format Transformation.}
    \label{fig:table_demo_2}
\end{figure*}
\begin{figure*}
    \centering
    \includegraphics[width=\linewidth]{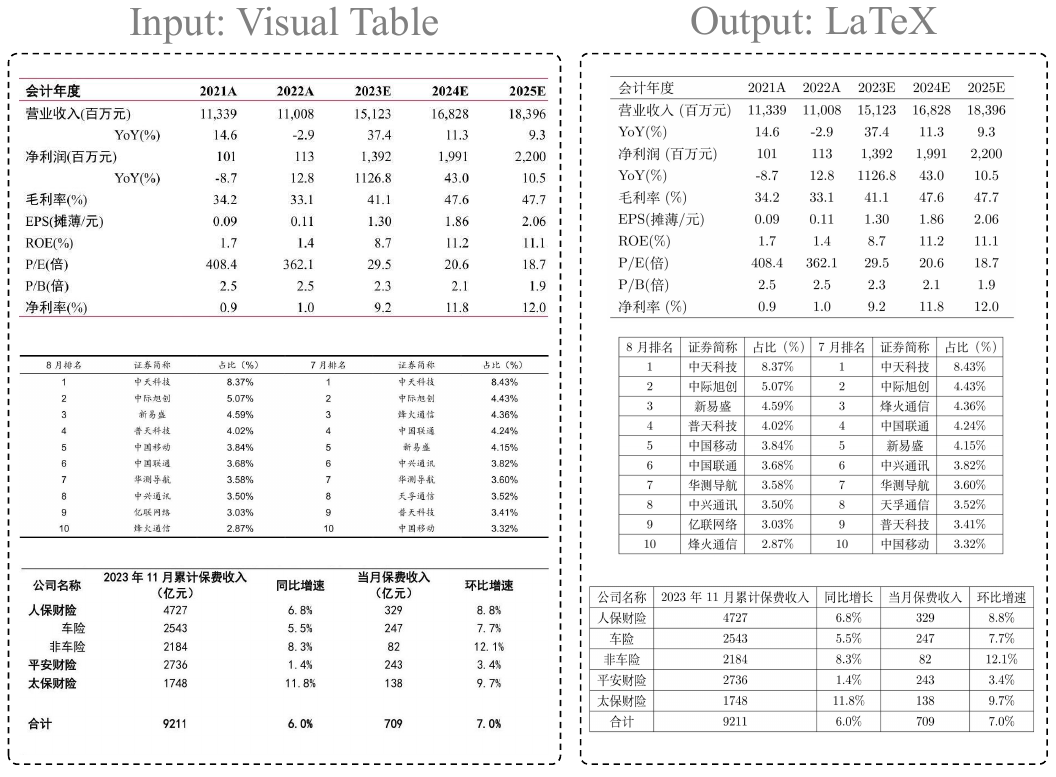}
    \caption{Output of Chimera-8B on Table Format Transformation.}
    \label{fig:table_demo_3}
\end{figure*}

\subsection{Chart Structural Extraction}
We provide the rendered table of the output results of Chimera-8B to show its chart structural extraction performance. 
As shown in Fig.~\ref{fig:chart_demo_1}, Fig.~\ref{fig:chart_demo_2} and Fig.~\ref{fig.:chart_demo_3}, 
Chimera can identify and extract information from various types of charts, such as pie charts, line graphs, bar charts, etc., and output this information in a structured format accurately.

\begin{figure*}
    \centering
    \includegraphics[width=\linewidth]{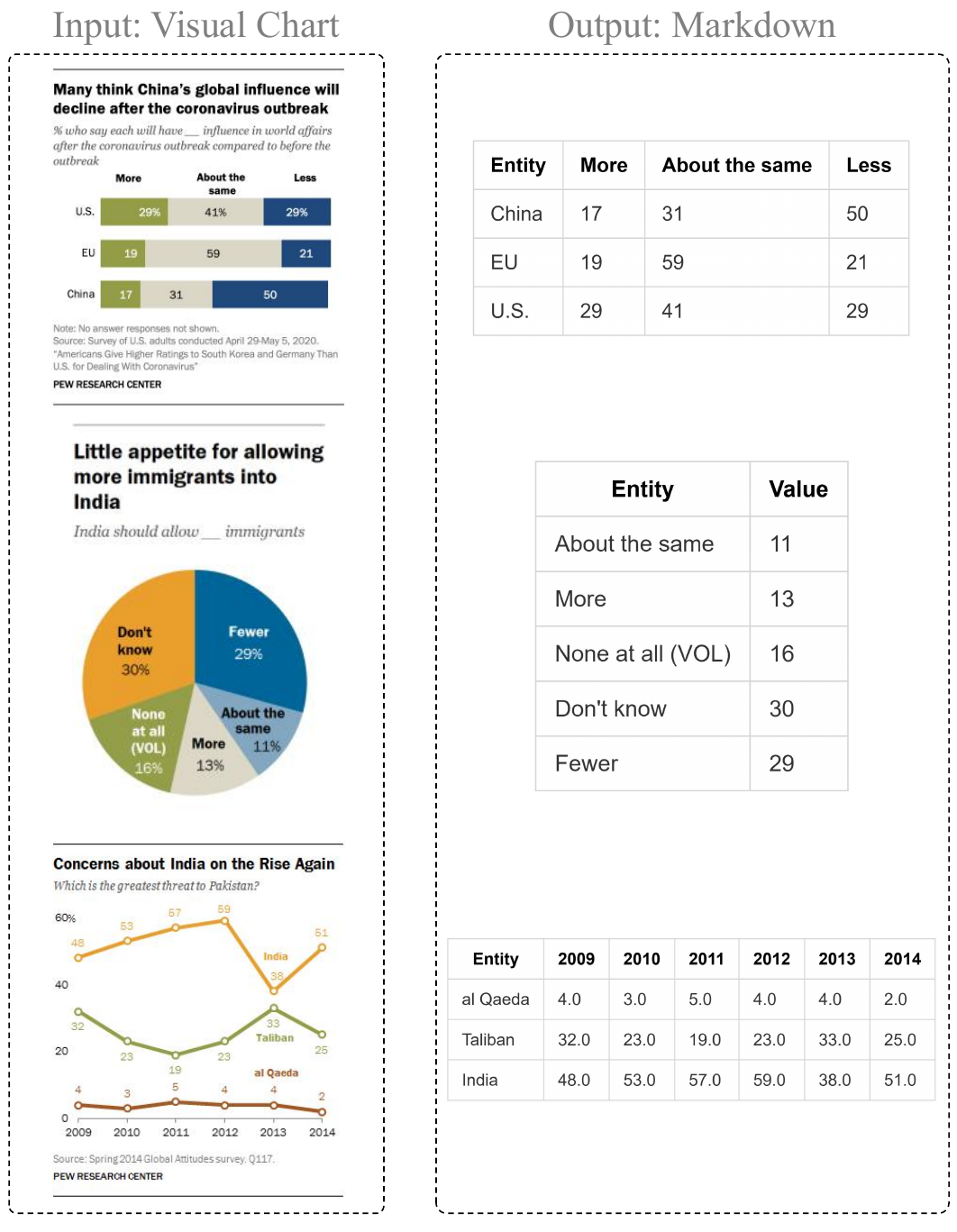}
    \caption{Output of Chimera-8B on Chart Structural Extraction.}
    \label{fig:chart_demo_1}
\end{figure*}
\begin{figure*}
    \centering
    \includegraphics[width=\linewidth]{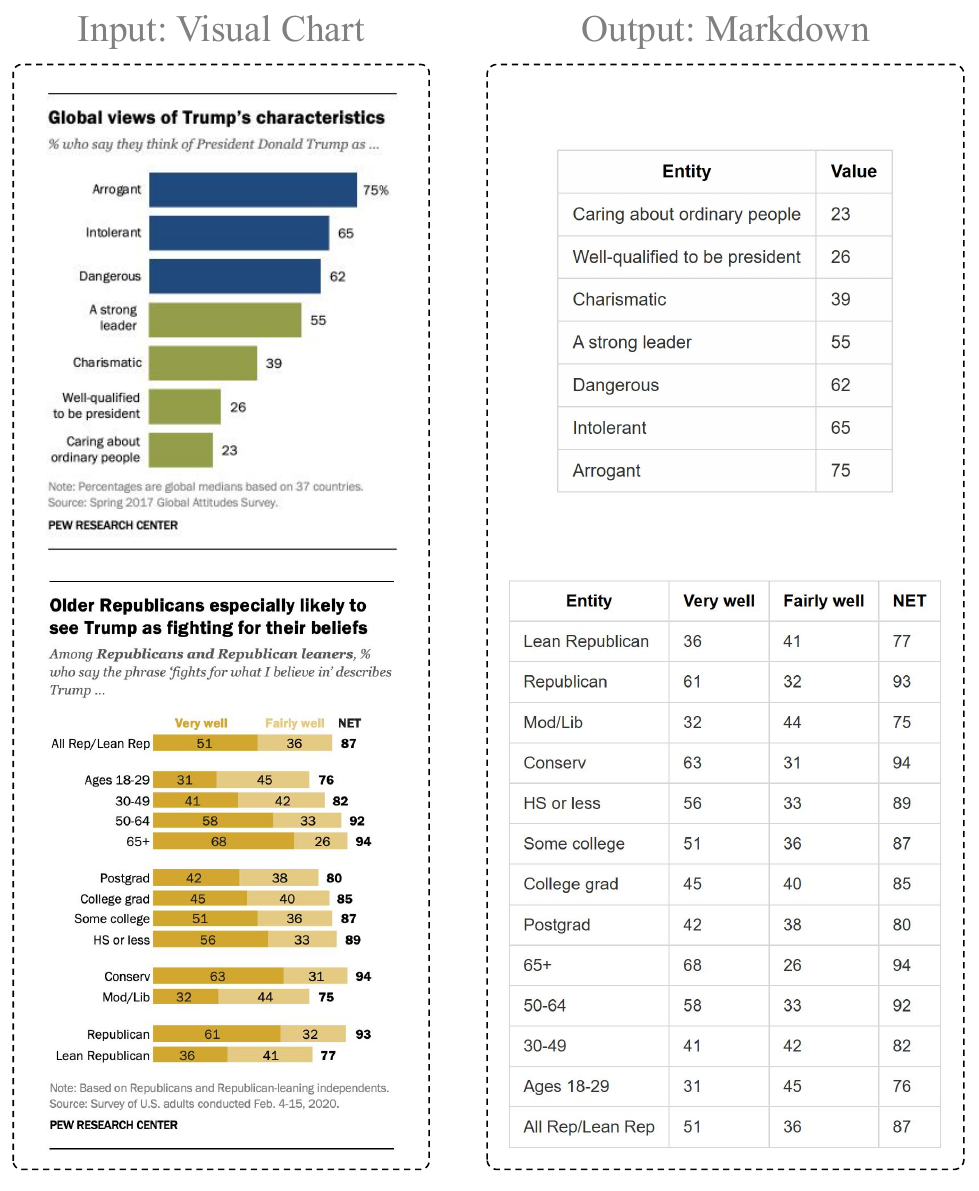}
    \caption{Output of Chimera-8B on Chart Structural Extraction.}
    \label{fig:chart_demo_2}
\end{figure*}
\begin{figure*}
    \centering
    \includegraphics[width=0.85\linewidth]{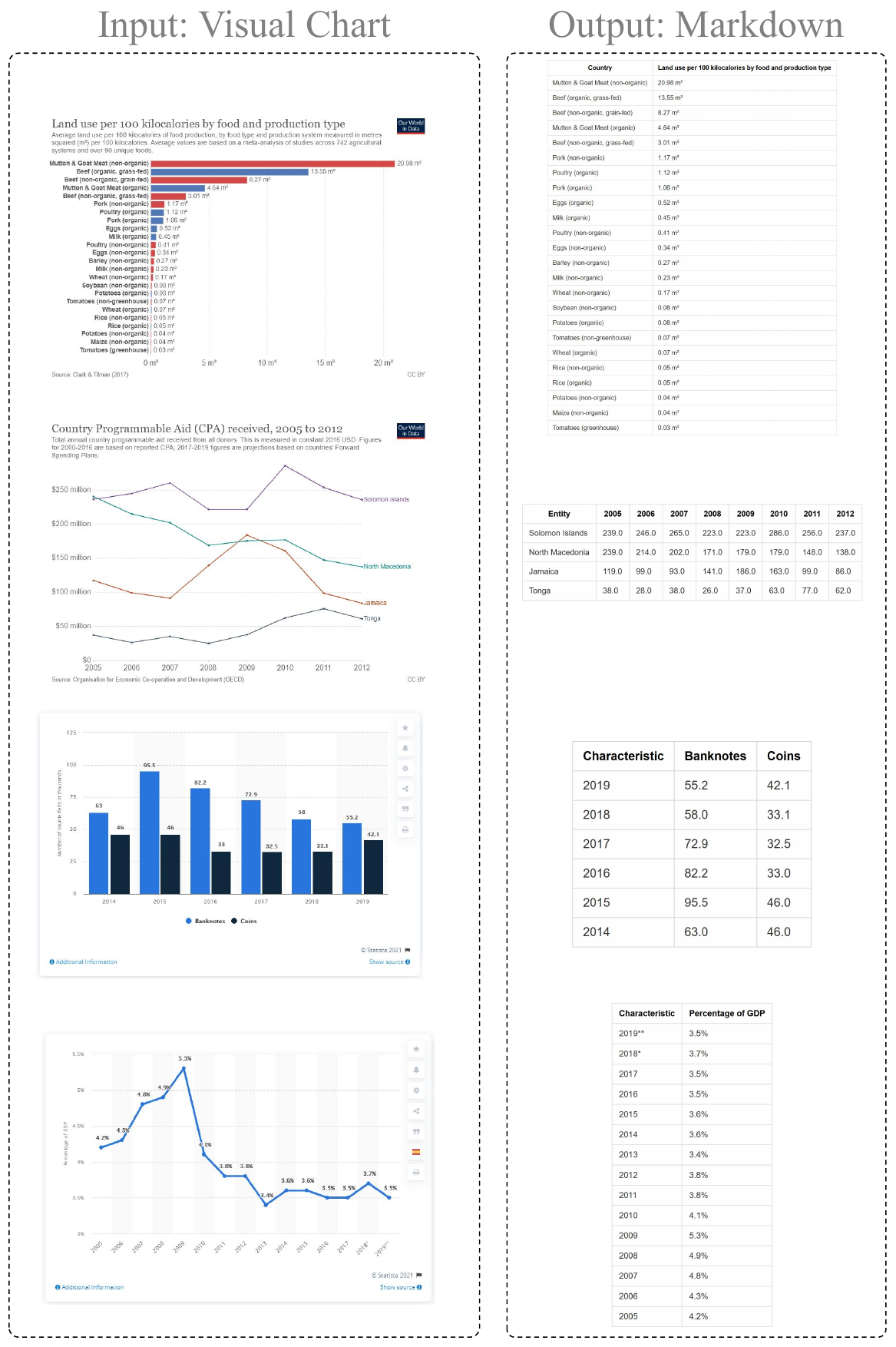}
    \caption{Output of Chimera-8B on Chart Structural Extraction.}
    \label{fig:chart_demo_3}
\end{figure*}

\subsection{Document Context Extraction}
We provide the rendered page of the output results of Chimera to show its document content extraction performance. 
As shown in Fig.~\ref{fig:page_demo_1}, Fig.~\ref{fig:page_demo_2}, Fig.~\ref{fig:page_demo_3} and Fig.~\ref{fig:page_demo_4}, Chimera demonstrates exceptional content extraction capabilities on both single-column and double-column documents, effectively extracting structured information end-to-end from text-dense visual inputs.

\begin{figure*}
    \centering
    \includegraphics[width=\linewidth]{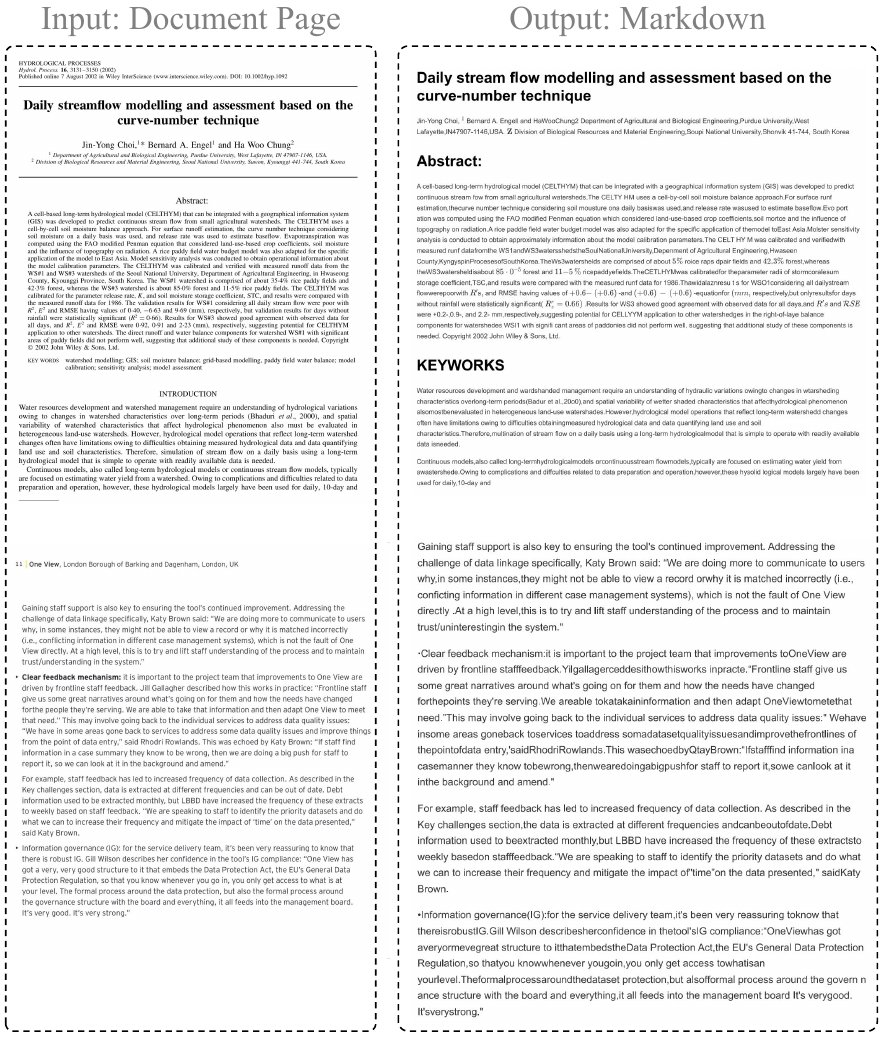}
    \caption{Output of Chimera on Document Context Extraction.}
    \label{fig:page_demo_1}
\end{figure*}
\begin{figure*}
    \centering
    \includegraphics[width=\linewidth]{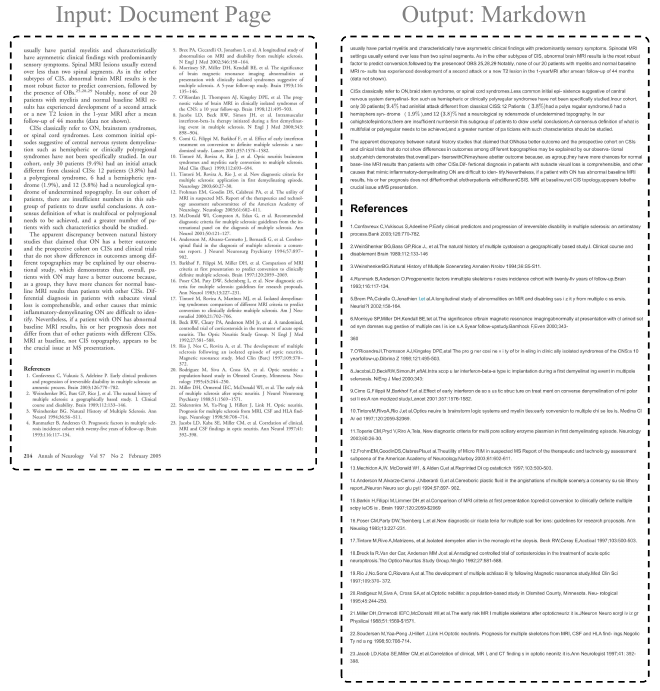}
    \caption{Output of Chimera} on Document Context Extraction.
    \label{fig:page_demo_2}
\end{figure*}
\begin{figure*}
    \centering
    \includegraphics[width=\linewidth]{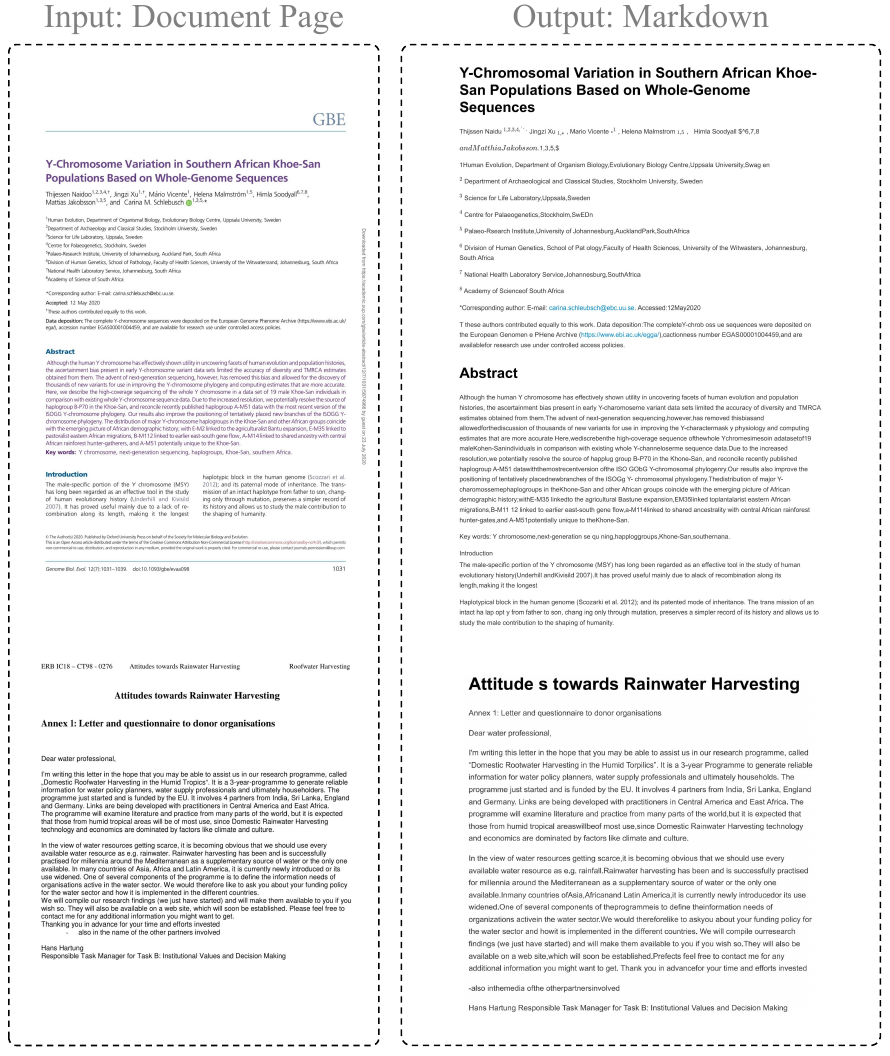}
    \caption{Output of Chimera on Document Context Extraction.}
    \label{fig:page_demo_3}
\end{figure*}
\begin{figure*}
    \centering
    \includegraphics[width=\linewidth]{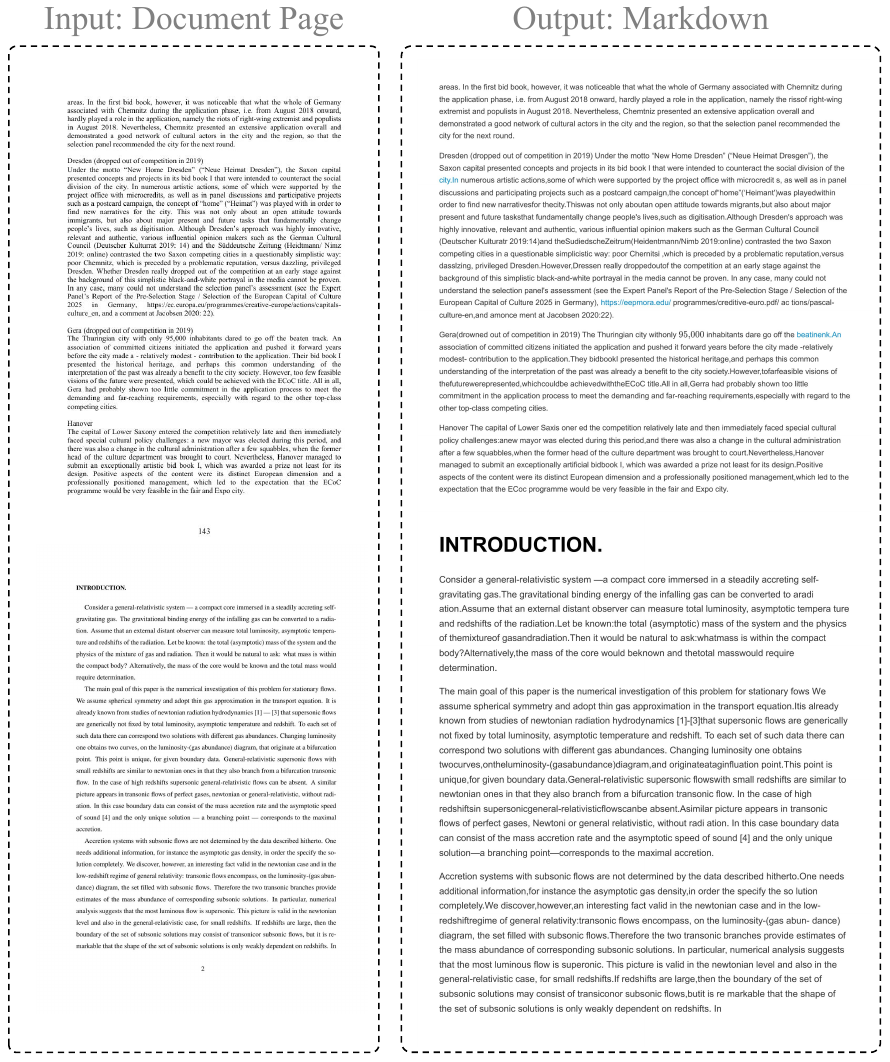}
    \caption{Output of Chimera on Document Context Extraction.}
    \label{fig:page_demo_4}
\end{figure*}

\clearpage

\clearpage
{
    \small
    \bibliographystyle{ieeenat_fullname}
    \bibliography{main}

\begin{thebibliography}{94}
\providecommand{\natexlab}[1]{#1}
\providecommand{\url}[1]{\texttt{#1}}
\expandafter\ifx\csname urlstyle\endcsname\relax
  \providecommand{\doi}[1]{doi: #1}\else
  \providecommand{\doi}{doi: \begingroup \urlstyle{rm}\Url}\fi

\bibitem[cla()]{claude3}
The claude 3 model family: Opus, sonnet, haiku.

\bibitem[Agrawal et~al.(2024)Agrawal, Antoniak, Hanna, Chaplot, Chudnovsky, Garg, Gervet, Ghosh, H{\'e}liou, Jacob, et~al.]{agrawal2024pixtral}
Pravesh Agrawal, Szymon Antoniak, Emma~Bou Hanna, Devendra Chaplot, Jessica Chudnovsky, Saurabh Garg, Theophile Gervet, Soham Ghosh, Am{\'e}lie H{\'e}liou, Paul Jacob, et~al.
\newblock Pixtral 12b.
\newblock \emph{arXiv preprint arXiv:2410.07073}, 2024.

\bibitem[Amini et~al.(2019)Amini, Gabriel, Lin, Koncel-Kedziorski, Choi, and Hajishirzi]{amini2019mathqa}
Aida Amini, Saadia Gabriel, Peter Lin, Rik Koncel-Kedziorski, Yejin Choi, and Hannaneh Hajishirzi.
\newblock Mathqa: Towards interpretable math word problem solving with operation-based formalisms.
\newblock \emph{arXiv preprint arXiv:1905.13319}, 2019.

\bibitem[Anthropic(2024)]{2024claude}
Anthropic.
\newblock The claude 3 model family: Opus, sonnet, haiku.
\newblock \url{https://www.anthropic.com,}, 2024.

\bibitem[Bai et~al.(2023)Bai, Bai, Yang, Wang, Tan, Wang, Lin, Zhou, and Zhou]{qwen-vl}
Jinze Bai, Shuai Bai, Shusheng Yang, Shijie Wang, Sinan Tan, Peng Wang, Junyang Lin, Chang Zhou, and Jingren Zhou.
\newblock Qwen-vl: A versatile vision-language model for understanding, localization, text reading, and beyond.
\newblock \emph{arXiv preprint arXiv:2308.12966}, 1\penalty0 (2):\penalty0 3, 2023.

\bibitem[Brown(2020)]{gpt3/brown2020language}
Tom~B Brown.
\newblock Language models are few-shot learners.
\newblock \emph{arXiv preprint ArXiv:2005.14165}, 2020.

\bibitem[Caffagni et~al.(2024)Caffagni, Cocchi, Moratelli, Sarto, Cornia, Baraldi, and Cucchiara]{caffagni2024wiki}
Davide Caffagni, Federico Cocchi, Nicholas Moratelli, Sara Sarto, Marcella Cornia, Lorenzo Baraldi, and Rita Cucchiara.
\newblock Wiki-llava: Hierarchical retrieval-augmented generation for multimodal llms.
\newblock In \emph{Proceedings of the IEEE/CVF Conference on Computer Vision and Pattern Recognition}, pages 1818--1826, 2024.

\bibitem[Chang et~al.(2022)Chang, Palzer, Li, Fosler-Lussier, and Xiao]{chang2022mapqa}
Shuaichen Chang, David Palzer, Jialin Li, Eric Fosler-Lussier, and Ningchuan Xiao.
\newblock Mapqa: A dataset for question answering on choropleth maps.
\newblock \emph{arXiv preprint arXiv:2211.08545}, 2022.

\bibitem[Chen et~al.(2021)Chen, Tang, Qin, Liang, Liu, Xing, and Lin]{chen2021geoqa}
Jiaqi Chen, Jianheng Tang, Jinghui Qin, Xiaodan Liang, Lingbo Liu, Eric~P Xing, and Liang Lin.
\newblock Geoqa: A geometric question answering benchmark towards multimodal numerical reasoning.
\newblock \emph{arXiv preprint arXiv:2105.14517}, 2021.

\bibitem[Chen et~al.(2023)Chen, Li, Dong, Zhang, He, Wang, Zhao, and Lin]{chen2023sharegpt4v}
Lin Chen, Jinsong Li, Xiaoyi Dong, Pan Zhang, Conghui He, Jiaqi Wang, Feng Zhao, and Dahua Lin.
\newblock Sharegpt4v: Improving large multi-modal models with better captions.
\newblock \emph{arXiv preprint arXiv:2311.12793}, 2023.

\bibitem[Chen et~al.(2024{\natexlab{a}})Chen, Chen, Zhang, Li, Yu, Fei, Zhu, Fan, and Chen]{ll3da}
Sijin Chen, Xin Chen, Chi Zhang, Mingsheng Li, Gang Yu, Hao Fei, Hongyuan Zhu, Jiayuan Fan, and Tao Chen.
\newblock Ll3da: Visual interactive instruction tuning for omni-3d understanding reasoning and planning.
\newblock In \emph{Proceedings of the IEEE/CVF Conference on Computer Vision and Pattern Recognition}, pages 26428--26438, 2024{\natexlab{a}}.

\bibitem[Chen et~al.(2024{\natexlab{b}})Chen, Wang, Tian, Ye, Gao, Cui, Tong, Hu, Luo, Ma, et~al.]{internvl-1.5}
Zhe Chen, Weiyun Wang, Hao Tian, Shenglong Ye, Zhangwei Gao, Erfei Cui, Wenwen Tong, Kongzhi Hu, Jiapeng Luo, Zheng Ma, et~al.
\newblock How far are we to gpt-4v? closing the gap to commercial multimodal models with open-source suites.
\newblock \emph{arXiv preprint arXiv:2404.16821}, 2024{\natexlab{b}}.

\bibitem[Chen et~al.(2024{\natexlab{c}})Chen, Wu, Wang, Su, Chen, Xing, Zhong, Zhang, Zhu, Lu, et~al.]{chen2024internvl}
Zhe Chen, Jiannan Wu, Wenhai Wang, Weijie Su, Guo Chen, Sen Xing, Muyan Zhong, Qinglong Zhang, Xizhou Zhu, Lewei Lu, et~al.
\newblock Internvl: Scaling up vision foundation models and aligning for generic visual-linguistic tasks.
\newblock In \emph{Proceedings of the IEEE/CVF Conference on Computer Vision and Pattern Recognition}, pages 24185--24198, 2024{\natexlab{c}}.

\bibitem[Dong et~al.(2024{\natexlab{a}})Dong, Li, Wu, Wang, Zhang, and Guo]{dong2024benchmarking}
Hongyuan Dong, Jiawen Li, Bohong Wu, Jiacong Wang, Yuan Zhang, and Haoyuan Guo.
\newblock Benchmarking and improving detail image caption.
\newblock \emph{arXiv preprint arXiv:2405.19092}, 2024{\natexlab{a}}.

\bibitem[Dong et~al.(2024{\natexlab{b}})Dong, Zhang, Zang, Cao, Wang, Ouyang, Wei, Zhang, Duan, Cao, et~al.]{ixc-2}
Xiaoyi Dong, Pan Zhang, Yuhang Zang, Yuhang Cao, Bin Wang, Linke Ouyang, Xilin Wei, Songyang Zhang, Haodong Duan, Maosong Cao, et~al.
\newblock Internlm-xcomposer2: Mastering free-form text-image composition and comprehension in vision-language large model.
\newblock \emph{arXiv preprint arXiv:2401.16420}, 2024{\natexlab{b}}.

\bibitem[Fu et~al.(2023)Fu, Chen, Shen, Qin, Zhang, Lin, Yang, Zheng, Li, Sun, et~al.]{fu2023mme}
Chaoyou Fu, Peixian Chen, Yunhang Shen, Yulei Qin, Mengdan Zhang, Xu Lin, Jinrui Yang, Xiawu Zheng, Ke Li, Xing Sun, et~al.
\newblock Mme: A comprehensive evaluation benchmark for multimodal large language models.
\newblock \emph{arXiv preprint arXiv:2306.13394}, 2023.

\bibitem[Gao et~al.(2023)Gao, Pi, Zhang, Ye, Zhong, Wang, Hong, Han, Xu, Li, et~al.]{gllava}
Jiahui Gao, Renjie Pi, Jipeng Zhang, Jiacheng Ye, Wanjun Zhong, Yufei Wang, Lanqing Hong, Jianhua Han, Hang Xu, Zhenguo Li, et~al.
\newblock G-llava: Solving geometric problem with multi-modal large language model.
\newblock \emph{arXiv preprint arXiv:2312.11370}, 2023.

\bibitem[He et~al.(2016)He, Zhang, Ren, and Sun]{he2016deep}
Kaiming He, Xiangyu Zhang, Shaoqing Ren, and Jian Sun.
\newblock Deep residual learning for image recognition.
\newblock In \emph{Proceedings of the IEEE conference on computer vision and pattern recognition}, pages 770--778, 2016.

\bibitem[Hu et~al.(2024)Hu, Xu, Zhang, Ye, Yan, Zhang, Jin, Huang, and Zhou]{hu2024mplug}
Anwen Hu, Haiyang Xu, Liang Zhang, Jiabo Ye, Ming Yan, Ji Zhang, Qin Jin, Fei Huang, and Jingren Zhou.
\newblock mplug-docowl2: High-resolution compressing for ocr-free multi-page document understanding.
\newblock \emph{arXiv preprint arXiv:2409.03420}, 2024.

\bibitem[Jiang et~al.(2025)Jiang, Zhang, Guo, Li, Qi, Chen, Wang, Jin, Guo, Yan, et~al.]{jiang2025mme}
Dongzhi Jiang, Renrui Zhang, Ziyu Guo, Yanwei Li, Yu Qi, Xinyan Chen, Liuhui Wang, Jianhan Jin, Claire Guo, Shen Yan, et~al.
\newblock Mme-cot: Benchmarking chain-of-thought in large multimodal models for reasoning quality, robustness, and efficiency.
\newblock \emph{arXiv preprint arXiv:2502.09621}, 2025.

\bibitem[Johnson et~al.(2017)Johnson, Hariharan, Van Der~Maaten, Fei-Fei, Lawrence~Zitnick, and Girshick]{johnson2017clevr}
Justin Johnson, Bharath Hariharan, Laurens Van Der~Maaten, Li Fei-Fei, C Lawrence~Zitnick, and Ross Girshick.
\newblock Clevr: A diagnostic dataset for compositional language and elementary visual reasoning.
\newblock In \emph{Proceedings of the IEEE conference on computer vision and pattern recognition}, pages 2901--2910, 2017.

\bibitem[Johnson et~al.(2019)Johnson, Douze, and J{\'e}gou]{johnson2019billion}
Jeff Johnson, Matthijs Douze, and Herv{\'e} J{\'e}gou.
\newblock Billion-scale similarity search with {GPUs}.
\newblock \emph{IEEE Transactions on Big Data}, 7\penalty0 (3):\penalty0 535--547, 2019.

\bibitem[Kafle et~al.(2018)Kafle, Price, Cohen, and Kanan]{kafle2018dvqa}
Kushal Kafle, Brian Price, Scott Cohen, and Christopher Kanan.
\newblock Dvqa: Understanding data visualizations via question answering.
\newblock In \emph{Proceedings of the IEEE conference on computer vision and pattern recognition}, pages 5648--5656, 2018.

\bibitem[Kahou et~al.(2017)Kahou, Michalski, Atkinson, K{\'a}d{\'a}r, Trischler, and Bengio]{kahou2017figureqa}
Samira~Ebrahimi Kahou, Vincent Michalski, Adam Atkinson, {\'A}kos K{\'a}d{\'a}r, Adam Trischler, and Yoshua Bengio.
\newblock Figureqa: An annotated figure dataset for visual reasoning.
\newblock \emph{arXiv preprint arXiv:1710.07300}, 2017.

\bibitem[Kantharaj et~al.(2022)Kantharaj, Leong, Lin, Masry, Thakkar, Hoque, and Joty]{kantharaj2022chart}
Shankar Kantharaj, Rixie Tiffany~Ko Leong, Xiang Lin, Ahmed Masry, Megh Thakkar, Enamul Hoque, and Shafiq Joty.
\newblock Chart-to-text: A large-scale benchmark for chart summarization.
\newblock \emph{arXiv preprint arXiv:2203.06486}, 2022.

\bibitem[Kazemi et~al.(2023)Kazemi, Alvari, Anand, Wu, Chen, and Soricut]{kazemi2023geomverse}
Mehran Kazemi, Hamidreza Alvari, Ankit Anand, Jialin Wu, Xi Chen, and Radu Soricut.
\newblock Geomverse: A systematic evaluation of large models for geometric reasoning.
\newblock \emph{arXiv preprint arXiv:2312.12241}, 2023.

\bibitem[Kembhavi et~al.(2016)Kembhavi, Salvato, Kolve, Seo, Hajishirzi, and Farhadi]{kembhavi2016diagram}
Aniruddha Kembhavi, Mike Salvato, Eric Kolve, Minjoon Seo, Hannaneh Hajishirzi, and Ali Farhadi.
\newblock A diagram is worth a dozen images.
\newblock In \emph{Computer Vision--ECCV 2016: 14th European Conference, Amsterdam, The Netherlands, October 11--14, 2016, Proceedings, Part IV 14}, pages 235--251. Springer, 2016.

\bibitem[Kembhavi et~al.(2017)Kembhavi, Seo, Schwenk, Choi, Farhadi, and Hajishirzi]{kembhavi2017you}
Aniruddha Kembhavi, Minjoon Seo, Dustin Schwenk, Jonghyun Choi, Ali Farhadi, and Hannaneh Hajishirzi.
\newblock Are you smarter than a sixth grader? textbook question answering for multimodal machine comprehension.
\newblock In \emph{Proceedings of the IEEE Conference on Computer Vision and Pattern recognition}, pages 4999--5007, 2017.

\bibitem[Li et~al.(2024{\natexlab{a}})Li, Zhang, Zhang, Guo, Zhang, Li, Zhang, Liu, and Li]{li2024llavanext-strong}
Bo Li, Kaichen Zhang, Hao Zhang, Dong Guo, Renrui Zhang, Feng Li, Yuanhan Zhang, Ziwei Liu, and Chunyuan Li.
\newblock Llava-next: Stronger llms supercharge multimodal capabilities in the wild, 2024{\natexlab{a}}.

\bibitem[Li et~al.(2024{\natexlab{b}})Li, Zhang, Guo, Zhang, Li, Zhang, Zhang, Li, Liu, and Li]{llava-onevision}
Bo Li, Yuanhan Zhang, Dong Guo, Renrui Zhang, Feng Li, Hao Zhang, Kaichen Zhang, Yanwei Li, Ziwei Liu, and Chunyuan Li.
\newblock Llava-onevision: Easy visual task transfer.
\newblock \emph{arXiv preprint arXiv:2408.03326}, 2024{\natexlab{b}}.

\bibitem[Li et~al.(2022)Li, Li, Xiong, and Hoi]{blip}
Junnan Li, Dongxu Li, Caiming Xiong, and Steven Hoi.
\newblock Blip: Bootstrapping language-image pre-training for unified vision-language understanding and generation.
\newblock In \emph{International conference on machine learning}, pages 12888--12900. PMLR, 2022.

\bibitem[Li et~al.(2023{\natexlab{a}})Li, Li, Savarese, and Hoi]{blip-2}
Junnan Li, Dongxu Li, Silvio Savarese, and Steven Hoi.
\newblock Blip-2: Bootstrapping language-image pre-training with frozen image encoders and large language models.
\newblock In \emph{International conference on machine learning}, pages 19730--19742. PMLR, 2023{\natexlab{a}}.

\bibitem[Li et~al.(2024{\natexlab{c}})Li, Wang, Xu, Wang, Feng, Kong, and Liu]{li2024multimodal}
Lei Li, Yuqi Wang, Runxin Xu, Peiyi Wang, Xiachong Feng, Lingpeng Kong, and Qi Liu.
\newblock Multimodal arxiv: A dataset for improving scientific comprehension of large vision-language models.
\newblock \emph{arXiv preprint arXiv:2403.00231}, 2024{\natexlab{c}}.

\bibitem[Li et~al.(2023{\natexlab{b}})Li, Chen, Zhang, Chen, Zhu, Yin, Yu, and Chen]{m3dbench}
Mingsheng Li, Xin Chen, Chi Zhang, Sijin Chen, Hongyuan Zhu, Fukun Yin, Gang Yu, and Tao Chen.
\newblock M3dbench: Let's instruct large models with multi-modal 3d prompts.
\newblock \emph{arXiv preprint arXiv:2312.10763}, 2023{\natexlab{b}}.

\bibitem[Li et~al.(2024{\natexlab{d}})Li, Chen, Wang, Wang, Ye, Jin, Chen, He, Gao, Cui, et~al.]{li2024omnicorpus}
Qingyun Li, Zhe Chen, Weiyun Wang, Wenhai Wang, Shenglong Ye, Zhenjiang Jin, Guanzhou Chen, Yinan He, Zhangwei Gao, Erfei Cui, et~al.
\newblock Omnicorpus: An unified multimodal corpus of 10 billion-level images interleaved with text.
\newblock \emph{arXiv preprint arXiv:2406.08418}, 2024{\natexlab{d}}.

\bibitem[Li et~al.(2023{\natexlab{c}})Li, Wang, Stengel-Eskin, Kortylewski, Ma, Van~Durme, and Yuille]{li2023super}
Zhuowan Li, Xingrui Wang, Elias Stengel-Eskin, Adam Kortylewski, Wufei Ma, Benjamin Van~Durme, and Alan~L Yuille.
\newblock Super-clevr: A virtual benchmark to diagnose domain robustness in visual reasoning.
\newblock In \emph{Proceedings of the IEEE/CVF Conference on Computer Vision and Pattern Recognition}, pages 14963--14973, 2023{\natexlab{c}}.

\bibitem[Liang et~al.(2022)Liang, Zhang, Kwon, Yeung, and Zou]{liang2022mind}
Victor~Weixin Liang, Yuhui Zhang, Yongchan Kwon, Serena Yeung, and James~Y Zou.
\newblock Mind the gap: Understanding the modality gap in multi-modal contrastive representation learning.
\newblock \emph{Advances in Neural Information Processing Systems}, 35:\penalty0 17612--17625, 2022.

\bibitem[Lifferth et~al.(2023)Lifferth, Reade, and Howard]{kaggle-llm-science-exam}
Will Lifferth, Walter Reade, and Addison Howard.
\newblock Kaggle - llm science exam.
\newblock \url{https://kaggle.com/competitions/kaggle-llm-science-exam}, 2023.
\newblock Kaggle.

\bibitem[Lin et~al.(2014)Lin, Maire, Belongie, Hays, Perona, Ramanan, Doll{\'a}r, and Zitnick]{coco}
Tsung-Yi Lin, Michael Maire, Serge Belongie, James Hays, Pietro Perona, Deva Ramanan, Piotr Doll{\'a}r, and C~Lawrence Zitnick.
\newblock Microsoft coco: Common objects in context.
\newblock In \emph{Computer Vision--ECCV 2014: 13th European Conference, Zurich, Switzerland, September 6-12, 2014, Proceedings, Part V 13}, pages 740--755. Springer, 2014.

\bibitem[Lin et~al.(2023)Lin, Liu, Zhang, Gao, Qiu, Xiao, Qiu, Lin, Shao, Chen, et~al.]{sphinx}
Ziyi Lin, Chris Liu, Renrui Zhang, Peng Gao, Longtian Qiu, Han Xiao, Han Qiu, Chen Lin, Wenqi Shao, Keqin Chen, et~al.
\newblock Sphinx: The joint mixing of weights, tasks, and visual embeddings for multi-modal large language models.
\newblock \emph{arXiv preprint arXiv:2311.07575}, 2023.

\bibitem[Liu et~al.(2024{\natexlab{a}})Liu, Akhgari, Visheratin, Kamko, Xu, Shrirao, Souza, Doshi, and Li]{liu2024playground}
Bingchen Liu, Ehsan Akhgari, Alexander Visheratin, Aleks Kamko, Linmiao Xu, Shivam Shrirao, Joao Souza, Suhail Doshi, and Daiqing Li.
\newblock Playground v3: Improving text-to-image alignment with deep-fusion large language models.
\newblock \emph{arXiv preprint arXiv:2409.10695}, 2024{\natexlab{a}}.

\bibitem[Liu et~al.(2022)Liu, Eisenschlos, Piccinno, Krichene, Pang, Lee, Joshi, Chen, Collier, and Altun]{liu2022deplot}
Fangyu Liu, Julian~Martin Eisenschlos, Francesco Piccinno, Syrine Krichene, Chenxi Pang, Kenton Lee, Mandar Joshi, Wenhu Chen, Nigel Collier, and Yasemin Altun.
\newblock Deplot: One-shot visual language reasoning by plot-to-table translation, 2022.

\bibitem[Liu et~al.(2023)Liu, Lin, Li, Wang, Yacoob, and Wang]{liu2023aligning}
Fuxiao Liu, Kevin Lin, Linjie Li, Jianfeng Wang, Yaser Yacoob, and Lijuan Wang.
\newblock Aligning large multi-modal model with robust instruction tuning.
\newblock \emph{arXiv preprint arXiv:2306.14565}, 2023.

\bibitem[Liu et~al.(2024{\natexlab{b}})Liu, Li, Li, and Lee]{llava-1.5}
Haotian Liu, Chunyuan Li, Yuheng Li, and Yong~Jae Lee.
\newblock Improved baselines with visual instruction tuning.
\newblock In \emph{Proceedings of the IEEE/CVF Conference on Computer Vision and Pattern Recognition}, pages 26296--26306, 2024{\natexlab{b}}.

\bibitem[Liu et~al.(2024{\natexlab{c}})Liu, Li, Wu, and Lee]{llava-1}
Haotian Liu, Chunyuan Li, Qingyang Wu, and Yong~Jae Lee.
\newblock Visual instruction tuning.
\newblock \emph{Advances in neural information processing systems}, 36, 2024{\natexlab{c}}.

\bibitem[Lu et~al.(2021{\natexlab{a}})Lu, Gong, Jiang, Qiu, Huang, Liang, and Zhu]{lu2021inter}
Pan Lu, Ran Gong, Shibiao Jiang, Liang Qiu, Siyuan Huang, Xiaodan Liang, and Song-Chun Zhu.
\newblock Inter-gps: Interpretable geometry problem solving with formal language and symbolic reasoning.
\newblock In \emph{The 59th Annual Meeting of the Association for Computational Linguistics (ACL)}, 2021{\natexlab{a}}.

\bibitem[Lu et~al.(2021{\natexlab{b}})Lu, Qiu, Chen, Xia, Zhao, Zhang, Yu, Liang, and Zhu]{lu2021iconqa}
Pan Lu, Liang Qiu, Jiaqi Chen, Tony Xia, Yizhou Zhao, Wei Zhang, Zhou Yu, Xiaodan Liang, and Song-Chun Zhu.
\newblock Iconqa: A new benchmark for abstract diagram understanding and visual language reasoning.
\newblock \emph{arXiv preprint arXiv:2110.13214}, 2021{\natexlab{b}}.

\bibitem[Lu et~al.(2023{\natexlab{a}})Lu, Bansal, Xia, Liu, Li, Hajishirzi, Cheng, Chang, Galley, and Gao]{lu2023mathvista}
Pan Lu, Hritik Bansal, Tony Xia, Jiacheng Liu, Chunyuan Li, Hannaneh Hajishirzi, Hao Cheng, Kai-Wei Chang, Michel Galley, and Jianfeng Gao.
\newblock Mathvista: Evaluating mathematical reasoning of foundation models in visual contexts.
\newblock \emph{arXiv preprint arXiv:2310.02255}, 2023{\natexlab{a}}.

\bibitem[Lu et~al.(2023{\natexlab{b}})Lu, Qiu, Chang, Wu, Zhu, Rajpurohit, Clark, and Kalyan]{lu2023dynamic}
Pan Lu, Liang Qiu, Kai-Wei Chang, Ying~Nian Wu, Song-Chun Zhu, Tanmay Rajpurohit, Peter Clark, and Ashwin Kalyan.
\newblock Dynamic prompt learning via policy gradient for semi-structured mathematical reasoning.
\newblock In \emph{International Conference on Learning Representations (ICLR)}, 2023{\natexlab{b}}.

\bibitem[Lu et~al.(2024)Lu, Li, Chen, Xu, Luo, Zhang, and Ye]{lu2024ovis}
Shiyin Lu, Yang Li, Qing-Guo Chen, Zhao Xu, Weihua Luo, Kaifu Zhang, and Han-Jia Ye.
\newblock Ovis: Structural embedding alignment for multimodal large language model.
\newblock \emph{arXiv:2405.20797}, 2024.

\bibitem[Lu et~al.(2025)Lu, Yuan, Li, Zhao, Qin, Li, Zhuo, Wen, Liu, Cao, et~al.]{lu2025omnicaptioner}
Yiting Lu, Jiakang Yuan, Zhen Li, Shitian Zhao, Qi Qin, Xinyue Li, Le Zhuo, Licheng Wen, Dongyang Liu, Yuewen Cao, et~al.
\newblock Omnicaptioner: One captioner to rule them all.
\newblock \emph{arXiv preprint arXiv:2504.07089}, 2025.

\bibitem[Masry et~al.(2022)Masry, Long, Tan, Joty, and Hoque]{masry2022chartqa}
Ahmed Masry, Do~Xuan Long, Jia~Qing Tan, Shafiq Joty, and Enamul Hoque.
\newblock Chartqa: A benchmark for question answering about charts with visual and logical reasoning.
\newblock \emph{arXiv preprint arXiv:2203.10244}, 2022.

\bibitem[Masry et~al.(2023)Masry, Kavehzadeh, Do, Hoque, and Joty]{masry2023unichart}
Ahmed Masry, Parsa Kavehzadeh, Xuan~Long Do, Enamul Hoque, and Shafiq Joty.
\newblock Unichart: A universal vision-language pretrained model for chart comprehension and reasoning.
\newblock \emph{arXiv preprint arXiv:2305.14761}, 2023.

\bibitem[Masry et~al.(2024{\natexlab{a}})Masry, Shahmohammadi, Parvez, Hoque, and Joty]{chartinstruct}
Ahmed Masry, Mehrad Shahmohammadi, Md~Rizwan Parvez, Enamul Hoque, and Shafiq Joty.
\newblock Chartinstruct: Instruction tuning for chart comprehension and reasoning.
\newblock \emph{arXiv preprint arXiv:2403.09028}, 2024{\natexlab{a}}.

\bibitem[Masry et~al.(2024{\natexlab{b}})Masry, Thakkar, Bajaj, Kartha, Hoque, and Joty]{chartgemma}
Ahmed Masry, Megh Thakkar, Aayush Bajaj, Aaryaman Kartha, Enamul Hoque, and Shafiq Joty.
\newblock Chartgemma: Visual instruction-tuning for chart reasoning in the wild.
\newblock \emph{arXiv preprint arXiv:2407.04172}, 2024{\natexlab{b}}.

\bibitem[Mathew et~al.(2021)Mathew, Karatzas, and Jawahar]{mathew2021docvqa}
Minesh Mathew, Dimosthenis Karatzas, and CV Jawahar.
\newblock Docvqa: A dataset for vqa on document images.
\newblock In \emph{Proceedings of the IEEE/CVF winter conference on applications of computer vision}, pages 2200--2209, 2021.

\bibitem[Methani et~al.(2020)Methani, Ganguly, Khapra, and Kumar]{methani2020plotqa}
Nitesh Methani, Pritha Ganguly, Mitesh~M Khapra, and Pratyush Kumar.
\newblock Plotqa: Reasoning over scientific plots.
\newblock In \emph{Proceedings of the IEEE/CVF Winter Conference on Applications of Computer Vision}, pages 1527--1536, 2020.

\bibitem[Mitra et~al.(2024)Mitra, Khanpour, Rosset, and Awadallah]{mitra2024orcamath}
Arindam Mitra, Hamed Khanpour, Corby Rosset, and Ahmed Awadallah.
\newblock Orca-math: Unlocking the potential of slms in grade school math, 2024.

\bibitem[OpenAI(2023)]{openai2023gpt4v}
OpenAI.
\newblock Gpt-4v.
\newblock \url{https://openai.com/index/gpt-4v-system-card/}, 2023.

\bibitem[OpenAI(2024)]{openai2024gpt4o}
OpenAI.
\newblock Hello gpt-4o.
\newblock \url{https://openai.com/index/hello-gpt-4o/}, 2024.

\bibitem[Ordonez et~al.(2011)Ordonez, Kulkarni, and Berg]{sbu}
Vicente Ordonez, Girish Kulkarni, and Tamara Berg.
\newblock Im2text: Describing images using 1 million captioned photographs.
\newblock \emph{Advances in neural information processing systems}, 24, 2011.

\bibitem[Reid et~al.(2024)Reid, Savinov, Teplyashin, Lepikhin, Lillicrap, Alayrac, Soricut, Lazaridou, Firat, Schrittwieser, et~al.]{reid2024gemini}
Machel Reid, Nikolay Savinov, Denis Teplyashin, Dmitry Lepikhin, Timothy Lillicrap, Jean-baptiste Alayrac, Radu Soricut, Angeliki Lazaridou, Orhan Firat, Julian Schrittwieser, et~al.
\newblock Gemini 1.5: Unlocking multimodal understanding across millions of tokens of context.
\newblock \emph{arXiv preprint arXiv:2403.05530}, 2024.

\bibitem[Saikh et~al.(2022)Saikh, Ghosal, Mittal, Ekbal, and Bhattacharyya]{saikh2022scienceqa}
Tanik Saikh, Tirthankar Ghosal, Amish Mittal, Asif Ekbal, and Pushpak Bhattacharyya.
\newblock Scienceqa: A novel resource for question answering on scholarly articles.
\newblock \emph{International Journal on Digital Libraries}, 23\penalty0 (3):\penalty0 289--301, 2022.

\bibitem[Schuhmann et~al.(2021)Schuhmann, Vencu, Beaumont, Kaczmarczyk, Mullis, Katta, Coombes, Jitsev, and Komatsuzaki]{laion}
Christoph Schuhmann, Richard Vencu, Romain Beaumont, Robert Kaczmarczyk, Clayton Mullis, Aarush Katta, Theo Coombes, Jenia Jitsev, and Aran Komatsuzaki.
\newblock Laion-400m: Open dataset of clip-filtered 400 million image-text pairs.
\newblock \emph{arXiv preprint arXiv:2111.02114}, 2021.

\bibitem[Sharma et~al.(2018)Sharma, Ding, Goodman, and Soricut]{cc3m}
Piyush Sharma, Nan Ding, Sebastian Goodman, and Radu Soricut.
\newblock Conceptual captions: A cleaned, hypernymed, image alt-text dataset for automatic image captioning.
\newblock In \emph{Proceedings of the 56th Annual Meeting of the Association for Computational Linguistics (Volume 1: Long Papers)}, pages 2556--2565, 2018.

\bibitem[Shen et~al.(2024)Shen, Chen, Shao, Guan, and Nie]{shen2024mome}
Leyang Shen, Gongwei Chen, Rui Shao, Weili Guan, and Liqiang Nie.
\newblock Mome: Mixture of multimodal experts for generalist multimodal large language models.
\newblock \emph{arXiv preprint arXiv:2407.12709}, 2024.

\bibitem[Shi et~al.(2024)Shi, Hu, Bin, Liu, Yang, Ng, Bing, and Lee]{mathllava}
Wenhao Shi, Zhiqiang Hu, Yi Bin, Junhua Liu, Yang Yang, See-Kiong Ng, Lidong Bing, and Roy Ka-Wei Lee.
\newblock Math-llava: Bootstrapping mathematical reasoning for multimodal large language models.
\newblock \emph{arXiv preprint arXiv:2406.17294}, 2024.

\bibitem[Tang et~al.(2023)Tang, Boggust, and Satyanarayan]{tang2023vistext}
Benny~J Tang, Angie Boggust, and Arvind Satyanarayan.
\newblock Vistext: A benchmark for semantically rich chart captioning.
\newblock \emph{arXiv preprint arXiv:2307.05356}, 2023.

\bibitem[Team et~al.(2023)Team, Anil, Borgeaud, Wu, Alayrac, Yu, Soricut, Schalkwyk, Dai, Hauth, et~al.]{team2023gemini}
Gemini Team, Rohan Anil, Sebastian Borgeaud, Yonghui Wu, Jean-Baptiste Alayrac, Jiahui Yu, Radu Soricut, Johan Schalkwyk, Andrew~M Dai, Anja Hauth, et~al.
\newblock Gemini: a family of highly capable multimodal models.
\newblock \emph{arXiv preprint arXiv:2312.11805}, 2023.

\bibitem[Team et~al.(2024)Team, Georgiev, Lei, Burnell, Bai, Gulati, Tanzer, Vincent, Pan, Wang, et~al.]{team2024gemini}
Gemini Team, Petko Georgiev, Ving~Ian Lei, Ryan Burnell, Libin Bai, Anmol Gulati, Garrett Tanzer, Damien Vincent, Zhufeng Pan, Shibo Wang, et~al.
\newblock Gemini 1.5: Unlocking multimodal understanding across millions of tokens of context.
\newblock \emph{arXiv preprint arXiv:2403.05530}, 2024.

\bibitem[Team et~al.(2025)Team, Zhang, Feng, Yan, Yuan, Yu, He, Huang, Hou, Nie, et~al.]{team2025novelseek}
NovelSeek Team, Bo Zhang, Shiyang Feng, Xiangchao Yan, Jiakang Yuan, Zhiyin Yu, Xiaohan He, Songtao Huang, Shaowei Hou, Zheng Nie, et~al.
\newblock Novelseek: When agent becomes the scientist--building closed-loop system from hypothesis to verification.
\newblock \emph{arXiv preprint arXiv:2505.16938}, 2025.

\bibitem[Tong et~al.(2024)Tong, Brown, Wu, Woo, Middepogu, Akula, Yang, Yang, Iyer, Pan, et~al.]{cambrian}
Shengbang Tong, Ellis Brown, Penghao Wu, Sanghyun Woo, Manoj Middepogu, Sai~Charitha Akula, Jihan Yang, Shusheng Yang, Adithya Iyer, Xichen Pan, et~al.
\newblock Cambrian-1: A fully open, vision-centric exploration of multimodal llms.
\newblock \emph{arXiv preprint arXiv:2406.16860}, 2024.

\bibitem[Touvron et~al.(2023{\natexlab{a}})Touvron, Lavril, Izacard, Martinet, Lachaux, Lacroix, Rozi{\`e}re, Goyal, Hambro, Azhar, et~al.]{llama/touvron2023llama}
Hugo Touvron, Thibaut Lavril, Gautier Izacard, Xavier Martinet, Marie-Anne Lachaux, Timoth{\'e}e Lacroix, Baptiste Rozi{\`e}re, Naman Goyal, Eric Hambro, Faisal Azhar, et~al.
\newblock Llama: Open and efficient foundation language models.
\newblock \emph{arXiv preprint arXiv:2302.13971}, 2023{\natexlab{a}}.

\bibitem[Touvron et~al.(2023{\natexlab{b}})Touvron, Martin, Stone, Albert, Almahairi, Babaei, Bashlykov, Batra, Bhargava, Bhosale, et~al.]{llama2/touvron2023llama}
Hugo Touvron, Louis Martin, Kevin Stone, Peter Albert, Amjad Almahairi, Yasmine Babaei, Nikolay Bashlykov, Soumya Batra, Prajjwal Bhargava, Shruti Bhosale, et~al.
\newblock Llama 2: Open foundation and fine-tuned chat models.
\newblock \emph{arXiv preprint arXiv:2307.09288}, 2023{\natexlab{b}}.

\bibitem[Wang et~al.(2024{\natexlab{a}})Wang, Xu, Zhao, Ouyang, Wu, Zhao, Xu, Liu, Qu, Shang, et~al.]{wang2024mineru}
Bin Wang, Chao Xu, Xiaomeng Zhao, Linke Ouyang, Fan Wu, Zhiyuan Zhao, Rui Xu, Kaiwen Liu, Yuan Qu, Fukai Shang, et~al.
\newblock Mineru: An open-source solution for precise document content extraction.
\newblock \emph{arXiv preprint arXiv:2409.18839}, 2024{\natexlab{a}}.

\bibitem[Wang et~al.(2024{\natexlab{b}})Wang, Pan, Shi, Lu, Zhan, and Li]{wang2024measuring}
Ke Wang, Junting Pan, Weikang Shi, Zimu Lu, Mingjie Zhan, and Hongsheng Li.
\newblock Measuring multimodal mathematical reasoning with math-vision dataset, 2024{\natexlab{b}}.

\bibitem[Wang et~al.(2024{\natexlab{c}})Wang, Bai, Tan, Wang, Fan, Bai, Chen, Liu, Wang, Ge, et~al.]{qwen2-vl}
Peng Wang, Shuai Bai, Sinan Tan, Shijie Wang, Zhihao Fan, Jinze Bai, Keqin Chen, Xuejing Liu, Jialin Wang, Wenbin Ge, et~al.
\newblock Qwen2-vl: Enhancing vision-language model's perception of the world at any resolution.
\newblock \emph{arXiv preprint arXiv:2409.12191}, 2024{\natexlab{c}}.

\bibitem[Wang et~al.(2024{\natexlab{d}})Wang, Bai, Tan, Wang, Fan, Bai, Chen, Liu, Wang, Ge, et~al.]{wang2024qwen2}
Peng Wang, Shuai Bai, Sinan Tan, Shijie Wang, Zhihao Fan, Jinze Bai, Keqin Chen, Xuejing Liu, Jialin Wang, Wenbin Ge, et~al.
\newblock Qwen2-vl: Enhancing vision-language model's perception of the world at any resolution.
\newblock \emph{arXiv preprint arXiv:2409.12191}, 2024{\natexlab{d}}.

\bibitem[Wang et~al.(2024{\natexlab{e}})Wang, Chen, Wang, Cao, Liu, Gao, Zhu, Zhu, Lu, Qiao, and Dai]{wang2024mpo}
Weiyun Wang, Zhe Chen, Wenhai Wang, Yue Cao, Yangzhou Liu, Zhangwei Gao, Jinguo Zhu, Xizhou Zhu, Lewei Lu, Yu Qiao, and Jifeng Dai.
\newblock Enhancing the reasoning ability of multimodal large language models via mixed preference optimization.
\newblock \emph{arXiv preprint arXiv:2411.10442}, 2024{\natexlab{e}}.

\bibitem[Wei et~al.(2024)Wei, Liu, Chen, Wang, Kong, Xu, Ge, Zhao, Sun, Peng, et~al.]{got}
Haoran Wei, Chenglong Liu, Jinyue Chen, Jia Wang, Lingyu Kong, Yanming Xu, Zheng Ge, Liang Zhao, Jianjian Sun, Yuang Peng, et~al.
\newblock General ocr theory: Towards ocr-2.0 via a unified end-to-end model.
\newblock \emph{arXiv preprint arXiv:2409.01704}, 2024.

\bibitem[Xia et~al.(2023)Xia, Zhang, Peng, Ye, Yan, Ye, Shi, Qiao, and Yan]{xia2023structchart}
Renqiu Xia, Bo Zhang, Haoyang Peng, Hancheng Ye, Xiangchao Yan, Peng Ye, Botian Shi, Yu Qiao, and Junchi Yan.
\newblock Structchart: Perception, structuring, reasoning for visual chart understanding.
\newblock \emph{arXiv preprint arXiv:2309.11268}, 2023.

\bibitem[Xia et~al.(2024{\natexlab{a}})Xia, Li, Ye, Wu, Zhou, Yuan, Peng, Cai, Yan, Wang, et~al.]{xia2024geox}
Renqiu Xia, Mingsheng Li, Hancheng Ye, Wenjie Wu, Hongbin Zhou, Jiakang Yuan, Tianshuo Peng, Xinyu Cai, Xiangchao Yan, Bin Wang, et~al.
\newblock Geox: Geometric problem solving through unified formalized vision-language pre-training.
\newblock \emph{arXiv preprint arXiv:2412.11863}, 2024{\natexlab{a}}.

\bibitem[Xia et~al.(2024{\natexlab{b}})Xia, Mao, Yan, Zhou, Zhang, Peng, Pi, Fu, Wu, Ye, et~al.]{docgenome}
Renqiu Xia, Song Mao, Xiangchao Yan, Hongbin Zhou, Bo Zhang, Haoyang Peng, Jiahao Pi, Daocheng Fu, Wenjie Wu, Hancheng Ye, et~al.
\newblock Docgenome: An open large-scale scientific document benchmark for training and testing multi-modal large language models.
\newblock \emph{arXiv preprint arXiv:2406.11633}, 2024{\natexlab{b}}.

\bibitem[Xia et~al.(2024{\natexlab{c}})Xia, Zhang, Ye, Yan, Liu, Zhou, Chen, Dou, Shi, Yan, et~al.]{chartvlm}
Renqiu Xia, Bo Zhang, Hancheng Ye, Xiangchao Yan, Qi Liu, Hongbin Zhou, Zijun Chen, Min Dou, Botian Shi, Junchi Yan, et~al.
\newblock Chartx \& chartvlm: A versatile benchmark and foundation model for complicated chart reasoning.
\newblock \emph{arXiv preprint arXiv:2402.12185}, 2024{\natexlab{c}}.

\bibitem[Ye et~al.(2024)Ye, Xu, Ye, Yan, Hu, Liu, Qian, Zhang, and Huang]{mplugowl-2}
Qinghao Ye, Haiyang Xu, Jiabo Ye, Ming Yan, Anwen Hu, Haowei Liu, Qi Qian, Ji Zhang, and Fei Huang.
\newblock mplug-owl2: Revolutionizing multi-modal large language model with modality collaboration.
\newblock In \emph{Proceedings of the IEEE/CVF Conference on Computer Vision and Pattern Recognition}, pages 13040--13051, 2024.

\bibitem[Yuan et~al.(2025)Yuan, Peng, Jiang, Lu, Zhang, Feng, Fu, Chen, Bai, Zhang, et~al.]{mme-reasoning}
Jiakang Yuan, Tianshuo Peng, Yilei Jiang, Yiting Lu, Renrui Zhang, Kaituo Feng, Chaoyou Fu, Tao Chen, Lei Bai, Bo Zhang, et~al.
\newblock Mme-reasoning: A comprehensive benchmark for logical reasoning in mllms.
\newblock \emph{arXiv preprint arXiv:2505.21327}, 2025.

\bibitem[Yue et~al.(2023)Yue, Qu, Zhang, Fu, Huang, Sun, Su, and Chen]{yue2023mammoth}
Xiang Yue, Xingwei Qu, Ge Zhang, Yao Fu, Wenhao Huang, Huan Sun, Yu Su, and Wenhu Chen.
\newblock Mammoth: Building math generalist models through hybrid instruction tuning.
\newblock \emph{arXiv preprint arXiv:2309.05653}, 2023.

\bibitem[Zhang et~al.(2024{\natexlab{a}})Zhang, Hu, Zhoubian, Du, Yang, Wang, Yue, Dong, and Tang]{zhang2024sciglm}
Dan Zhang, Ziniu Hu, Sining Zhoubian, Zhengxiao Du, Kaiyu Yang, Zihan Wang, Yisong Yue, Yuxiao Dong, and Jie Tang.
\newblock Sciglm: Training scientific language models with self-reflective instruction annotation and tuning.
\newblock \emph{arXiv preprint arXiv:2401.07950}, 2024{\natexlab{a}}.

\bibitem[Zhang et~al.(2024{\natexlab{b}})Zhang, Wei, Jiang, Zhang, Guo, Tong, Liu, Zhou, Wei, Zhang, et~al.]{mavis}
Renrui Zhang, Xinyu Wei, Dongzhi Jiang, Yichi Zhang, Ziyu Guo, Chengzhuo Tong, Jiaming Liu, Aojun Zhou, Bin Wei, Shanghang Zhang, et~al.
\newblock Mavis: Mathematical visual instruction tuning.
\newblock \emph{arXiv preprint arXiv:2407.08739}, 2024{\natexlab{b}}.

\bibitem[Zhang et~al.(2025)Zhang, Jiang, Zhang, Lin, Guo, Qiu, Zhou, Lu, Chang, Qiao, et~al.]{zhang2025mathverse}
Renrui Zhang, Dongzhi Jiang, Yichi Zhang, Haokun Lin, Ziyu Guo, Pengshuo Qiu, Aojun Zhou, Pan Lu, Kai-Wei Chang, Yu Qiao, et~al.
\newblock Mathverse: Does your multi-modal llm truly see the diagrams in visual math problems?
\newblock In \emph{European Conference on Computer Vision}, pages 169--186. Springer, 2025.

\bibitem[Zhang et~al.(2023)Zhang, Zhang, Gu, Zhou, Lipka, Yang, and Sun]{zhang2023llavar}
Yanzhe Zhang, Ruiyi Zhang, Jiuxiang Gu, Yufan Zhou, Nedim Lipka, Diyi Yang, and Tong Sun.
\newblock Llavar: Enhanced visual instruction tuning for text-rich image understanding.
\newblock \emph{arXiv preprint arXiv:2306.17107}, 2023.

\bibitem[Zheng et~al.(2024)Zheng, Feng, Si, She, Lin, Jiang, and Wang]{tablellava}
Mingyu Zheng, Xinwei Feng, Qingyi Si, Qiaoqiao She, Zheng Lin, Wenbin Jiang, and Weiping Wang.
\newblock Multimodal table understanding.
\newblock \emph{arXiv preprint arXiv:2406.08100}, 2024.

\bibitem[Zhuang et~al.(2024)Zhuang, Huang, Zhang, and Zeng]{math-puma}
Wenwen Zhuang, Xin Huang, Xiantao Zhang, and Jin Zeng.
\newblock Math-puma: Progressive upward multimodal alignment to enhance mathematical reasoning.
\newblock \emph{arXiv preprint arXiv:2408.08640}, 2024.

\bibitem[Zong et~al.(2024)Zong, Ma, Shen, Song, Shao, Jiang, Li, and Liu]{zong2024mova}
Zhuofan Zong, Bingqi Ma, Dazhong Shen, Guanglu Song, Hao Shao, Dongzhi Jiang, Hongsheng Li, and Yu Liu.
\newblock Mova: Adapting mixture of vision experts to multimodal context.
\newblock \emph{arXiv preprint arXiv:2404.13046}, 2024.

\end{thebibliography}
}

\end{document}